%% file: 0_main.tex
\documentclass[runningheads]{llncs}
\usepackage{graphicx}
\usepackage{amsmath,amssymb} 
\usepackage{color}
\usepackage{booktabs}
\usepackage{array}
\usepackage{pifont}
\usepackage{hhline}
\usepackage{dsfont}
\usepackage{tabu}
\begin{document}
\newcolumntype{?}[1]{!{\vrule width #1}}
\pagestyle{headings}
\mainmatter
\def\ECCV18SubNumber{1567}  

\title{Expressive Telepresence via \\ Modular Codec Avatars} 

\titlerunning{Expressive Telepresence via Modular Codec Avatars}

\authorrunning{H. Chu, S. Ma, F. De la Torre, S. Fidler, and Y. Sheikh}

\author{Hang Chu$^{1,2}$ \hspace{50pt} Shugao Ma$^{3}$ \hspace{20pt} Fernando De la Torre$^{3}$ \\ Sanja Fidler$^{1,2}$ \hspace{35pt} Yaser Sheikh$^{3}$}
\institute{$^{1}$University of Toronto \qquad $^{2}$Vector Institute \qquad $^{3}$Facebook Reality Lab}

\maketitle

\begin{abstract}
VR telepresence consists of interacting with another human in a virtual space represented by an avatar. Today most avatars are cartoon-like, but soon the technology will allow video-realistic ones. This paper aims in this direction, and presents 
Modular Codec Avatars (MCA), a method to generate hyper-realistic faces driven by the cameras in the VR headset.
MCA extends traditional Codec Avatars (CA) by replacing the holistic models with a learned modular representation. It is important to note that traditional person-specific CAs are learned from few training samples, and typically lack robustness as well as limited expressiveness when transferring facial expressions. MCAs solve these issues by learning a modulated adaptive blending of different facial components as well as an exemplar-based latent alignment. We demonstrate that MCA achieves improved expressiveness and robustness w.r.t to CA in a variety of real-world datasets and practical scenarios. Finally, we showcase new applications in VR telepresence enabled by the proposed model.

\end{abstract}

\input{1_introduction.tex}
\input{2_related.tex}
\input{3_dataset.tex}
\input{4_method.tex}
\input{5_experiment.tex}
\input{6_conclusion.tex}

\clearpage

\bibliographystyle{splncs}
\bibliography{0_main}
\end{document}

%% file: 1_introduction.tex
\section{Introduction}
Telepresence technologies aims to make a person feel as if they were present, to give the appearance of being present, or to have an effect via telerobotics, at a place other than their true location. Telepresence systems can be broadly categorized based on the level of immersiveness. 
The most basic form of telepresence is video teleconferencing (e.g., Skype, Hangouts, Messenger) that is widely-used, and includes both audio and video transmissions.  Recently, a more sophisticated form of telepresence has become available featuring a 
smart camera that follows a person (e.g., the Portal from Facebook). 

\begin{figure}[t!]
    \centering
    \includegraphics[width=0.9\textwidth]{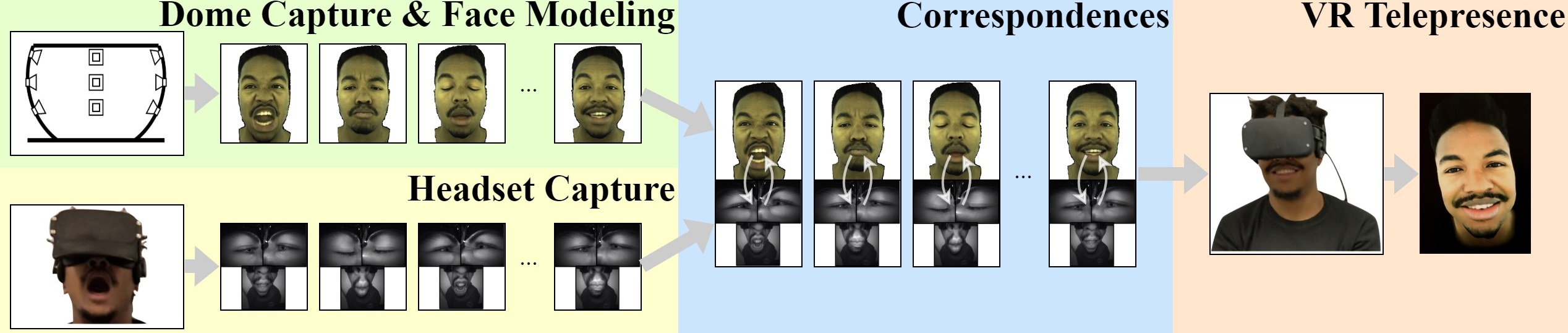}
    \caption{\footnotesize Train and test pipeline for our VR telepresence system. In the first stage, we capture facial expressions of a user using both a multi-view camera dome and a VR headset (mounted with face-looking cameras). Correspondences between VR headset recording and full face expressions are established using the method described in ~\cite{wei2019vr}. Finally, once the person-specific face animation model is learned using these correspondences, a real-time photo-realistic avatar is driven from the VR headset cameras.}
    \label{fig:overall}
\end{figure}

This paper addresses a more immersive form of telepresence that utilizes a virtual reality (VR) headset (e.g.,  Oculus, VIVE headsets). VR telepresence aims to enable a telecommunication system that allows remote social interactions more immersive than any prior media.  It is not only a key promise of VR, but also has vast potential socioeconomic impact such as increasing communication efficiency, lowering energy footprint, and such a system could be timely for reducing inter-personal disease transmission~\cite{heymann2020covid}. VR telepresence has been an important active area of research in computer vision~\cite{orts2016holoportation,wei2019vr,lombardi2018deep,tewari2019fml,thies2016face2face,elgharib2019egoface,nagano2018pagan}. In VR telepresence, 
the users wear the VR headset, and a 3D face avatar is holographically  projected in realtime, as if the user teleports himself/herself into the virtual space. 
This allows immersive bidirectional face-to-face conversations, facilitating instant interpersonal communication with high fidelity.

Fig.~\ref{fig:overall} illustrates our VR telepresence system that has three main stages: (1) Appearance/shape capture. The person-specific avatar is built by capturing shape and appearance of the person from a multi-camera system (i.e., the dome). The user performs the same set of scripted facial expressions siting in the dome and wearing a headset mounted with face-looking cameras respectively. The 3D faces are reconstructed from the dome-captured multi-view images, and a variational autoencoder (VAE) is trained to model the 3D shape and appearance variability of the person's face.
This model is referred to as Codec Avatar (CA)~\cite{lombardi2018deep,wei2019vr}, since it decodes the 3D face from low-dimensional code. The CA is learned from 10k-14k 3D shapes and texture images.   (2) Learning the 
correspondence between the infra-red (IR) VR cameras and the codec avatar. In the second stage, the CA method establishes the correspondence between the headset cameras and the 3D face model using a image-based synthesis approach~\cite{wei2019vr}. Once a set of IR images (i.e., mouth, eyes) in the VR headset are in correspondence with the CA, we learn a network to map the  IR VR headset cameras to the codec avatar codes, that should generalize to unseen situations (e.g. expressions, environments).  (3) Realtime inference. Given the input images from the VR headset, and the network learned in step two, we can drive a person-specific and photo-realistic face avatar. However, in this stage the CA has to satisfy two properties for authentic interactive VR experience:  
\begin{itemize}
    \item \textbf{Expressiveness}: The VR system needs to transfer the subtle expressions of the user. However, there are several challenges: (1) The CA model has been learned from limited training samples of the user ($\sim$10k-14k), and the system has to precisely interpolate/extrapolate unseen expressions. Recall that is impractical to have a uniform sample of all possible expressions in the training set, because of their long-tail distribution.  This requires careful ML algorithms that can learn from few training samples and long-tail distributions. (2) In addition, CA is an holistic model that typically results in rigid facial expression transfers. 
    \item \textbf{Robustness}: To enable VR telepresence at scale in realistic scenarios, CAs have to provide robustness across different sources of variability, that includes: (1) iconic changes in the users' appearance (e.g., beard, makeup), (2) variability in the headset camera position (e.g., head strap), (3) different lighting and background from different room environments, (4) hardware variations within manufacturing specification tolerance (e.g., LED intensity, camera placement). 
\end{itemize}

\begin{figure}[t!]
    \centering
    \setlength{\tabcolsep}{0pt}
    \begin{tabular}{c}
    \includegraphics[height=0.18\textwidth]{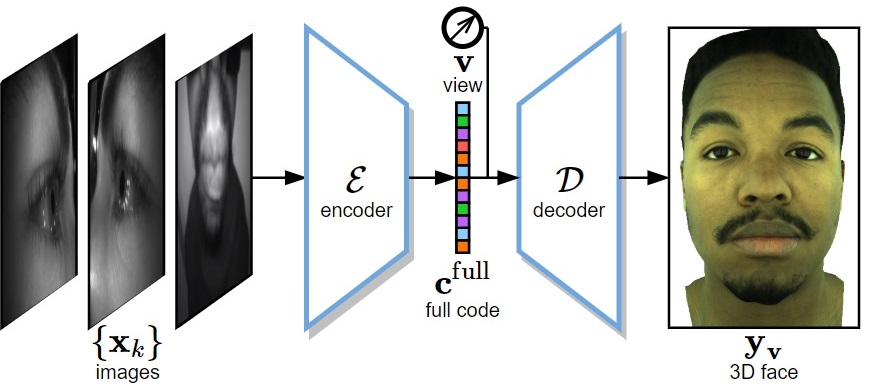}\\
    {\footnotesize Codec Avatar~\cite{wei2019vr,lombardi2018deep}}\\
    \includegraphics[height=0.18\textwidth]{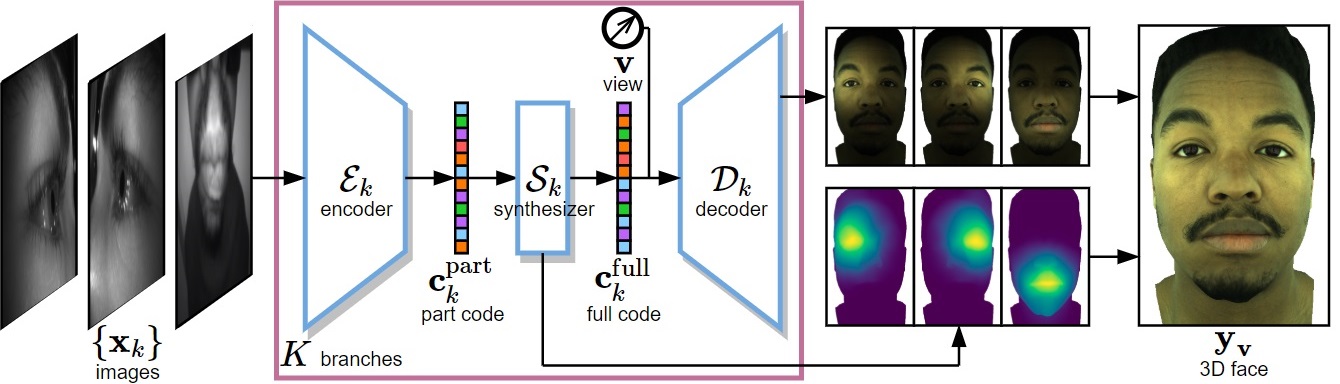}\\
    {\footnotesize Modular Codec Avatar (MCA)}\\
    \end{tabular}
    \caption{\footnotesize Model diagrams comparing the previous CA and the proposed MCA. $K$ denote the number of head-mounted cameras. In CA, images of all headset cameras are feed together to the single encoder $\mathcal{E}$ to compute the full face code which is subsequently decoded into 3D face using deocoder $\mathcal{D}$.  In MCA, the images of each camera are encoded separately into a modular code $\mathbf{c}^{\mathrm{part}}_k$ with the encoder $\mathcal{E}_k$, which is feed to a synthesizer $\mathcal{S}_k$ to estimate a camera specific full face code $\mathbf{c}_k^{\mathrm{full}}$, and blending weights. Finally all these camera specific full face codes are decoded into 3D faces and blended together to form the final face avatar.}
    \label{fig:algorithm}
\end{figure}

This paper proposes Modular Codec Avatar (MCA) that improves robustness and expressiveness of traditional CA. MCA decomposed the holistic face representation of traditional CA into learned modular local representations. Each local representation corresponds to one headset-mounted camera. MCA learns from data the automatic blending across all the modules.  Fig.~\ref{fig:algorithm} shows a diagram comparing the CA and MCA models. In MCA, a modular encoder first extracts information inside each single headset-mounted camera view. This is followed by a modular synthesizer that estimates a full face expression along with its blending weights from the information extracted within the same modular branch. Finally, multiple estimated 3D faces are aggregated from different modules and blended together to form the final face output.

Our contributions are threefold. First, we present MCA that introduces modularity into CA. Second, we extend MCA to solve the expressivity and robustness issues by learning the blending as well as new constraints in the latent space. Finally, we demostrate MCA's robustness and expressiveness advantages on a real-world VR dataset. 

%% file: 2_related.tex
\section{Related Work}
\newcolumntype{x}[1]{>{\centering\let\newline\\\arraybackslash\hspace{0pt}}p{#1}}
\begin{figure}[!t]
\centering
\setlength{\tabcolsep}{1pt}
\begin{tabular}{c?{.1em}l}
\specialrule{.1em}{.05em}{.05em}
    \rotatebox[origin=c]{90}{{\scriptsize hardware}} &
    \tabulinesep=\tabcolsep
    \begin{tabu} to 0.97\textwidth {X[0.7294,c,m]X[0.5183,c,m]X[0.4689,c,m]X[1.0,c,m]}
    \includegraphics[height=0.162\textwidth]{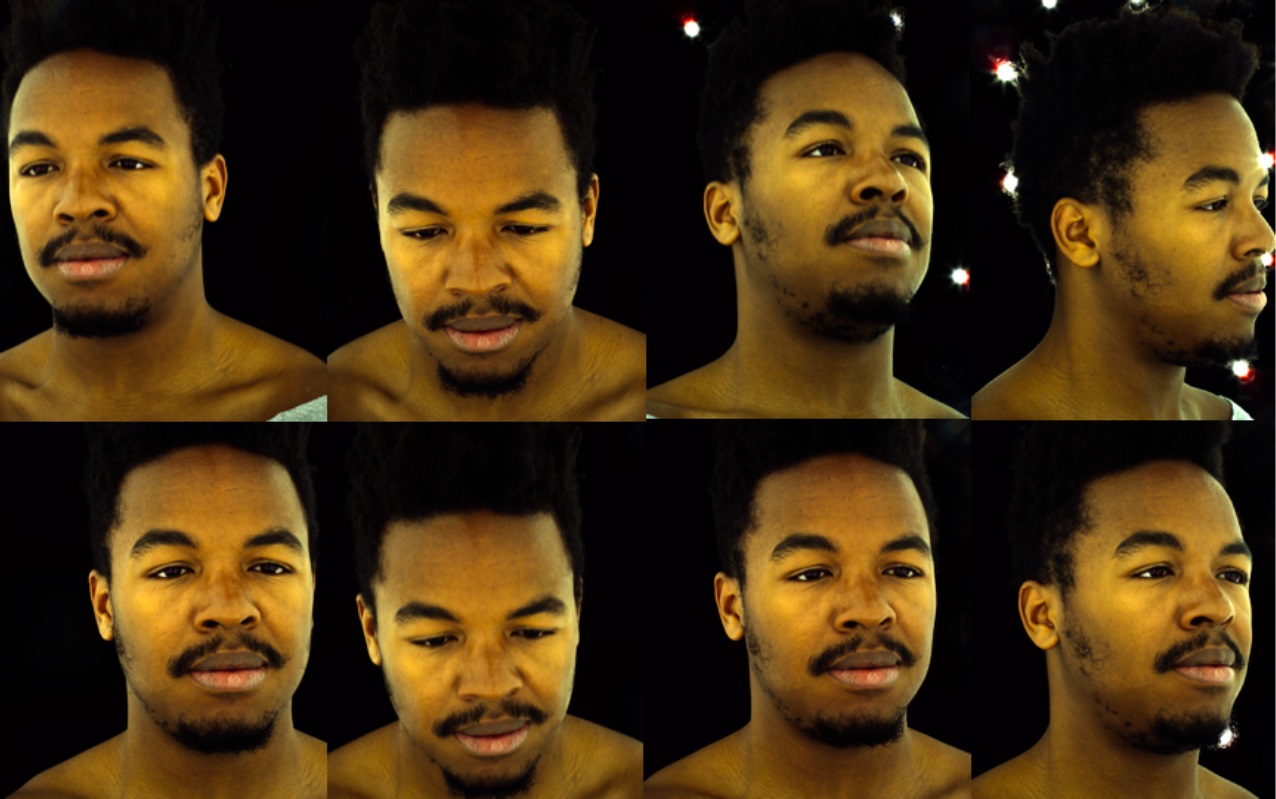} &
    \includegraphics[height=0.162\textwidth]{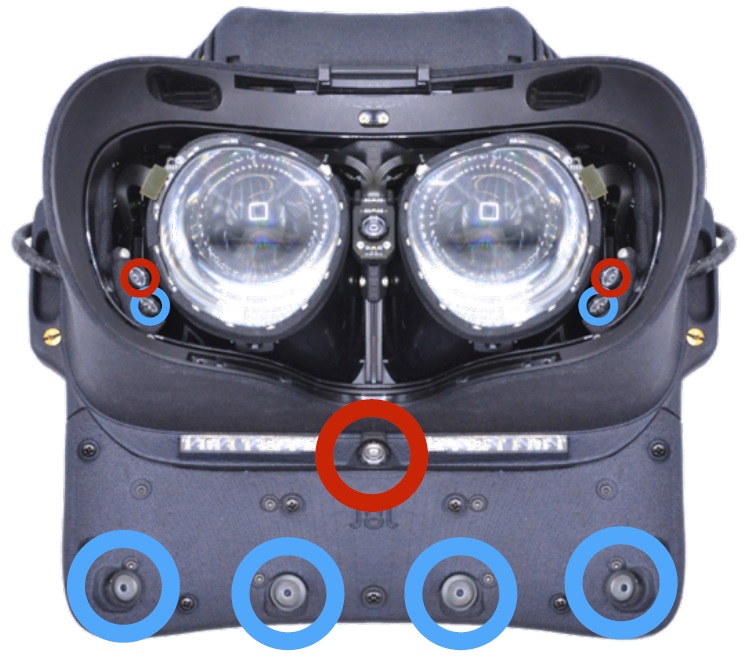} & 
    \includegraphics[height=0.162\textwidth]{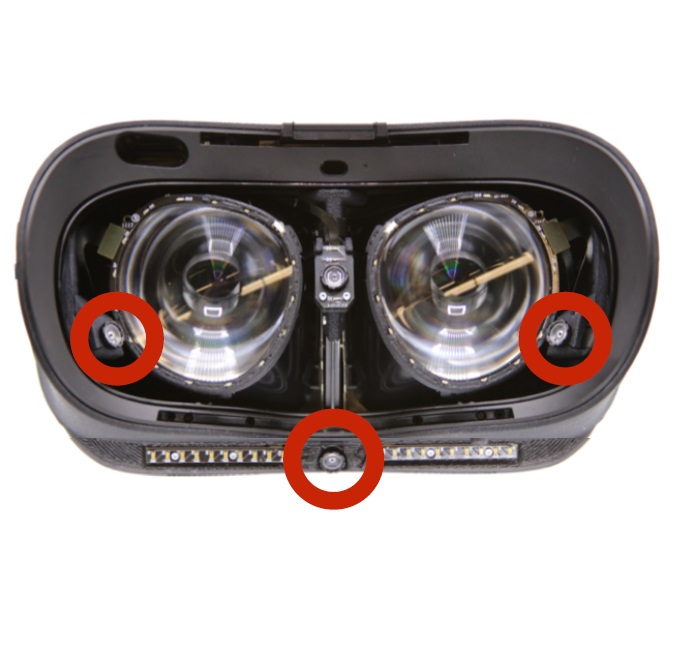} & 
    \includegraphics[height=0.162\textwidth]{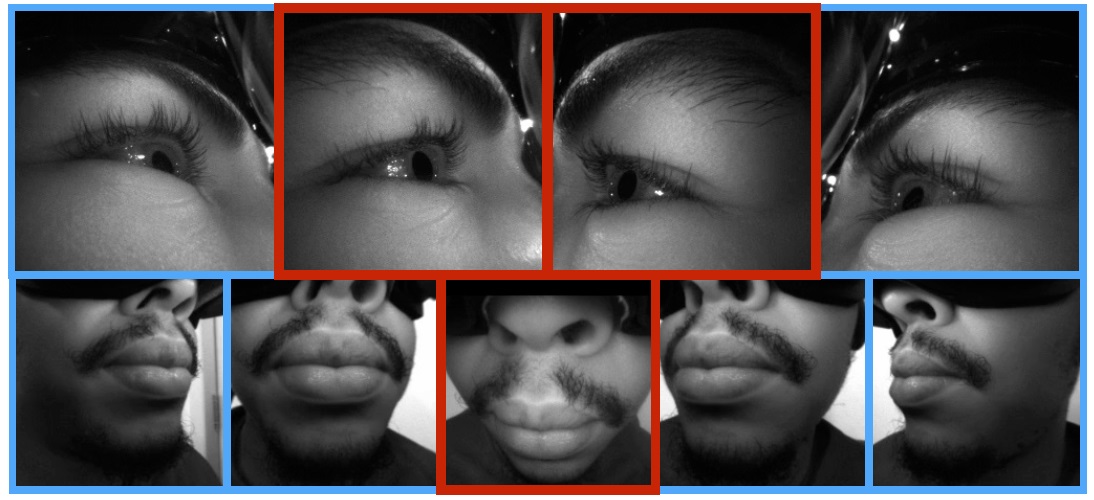}\\
    \end{tabu}
    \\\specialrule{.1em}{.05em}{.05em}
    \rotatebox[origin=c]{90}{{\scriptsize images}} &
    \tabulinesep=\tabcolsep
    \begin{tabu} to 0.97\textwidth {X[c,m]X[c,m]X[c,m]X[c,m]X[c,m]X[c,m]X[c,m]X[c,m]X[c,m]X[c,m]}
    \includegraphics[height=0.209\textwidth]{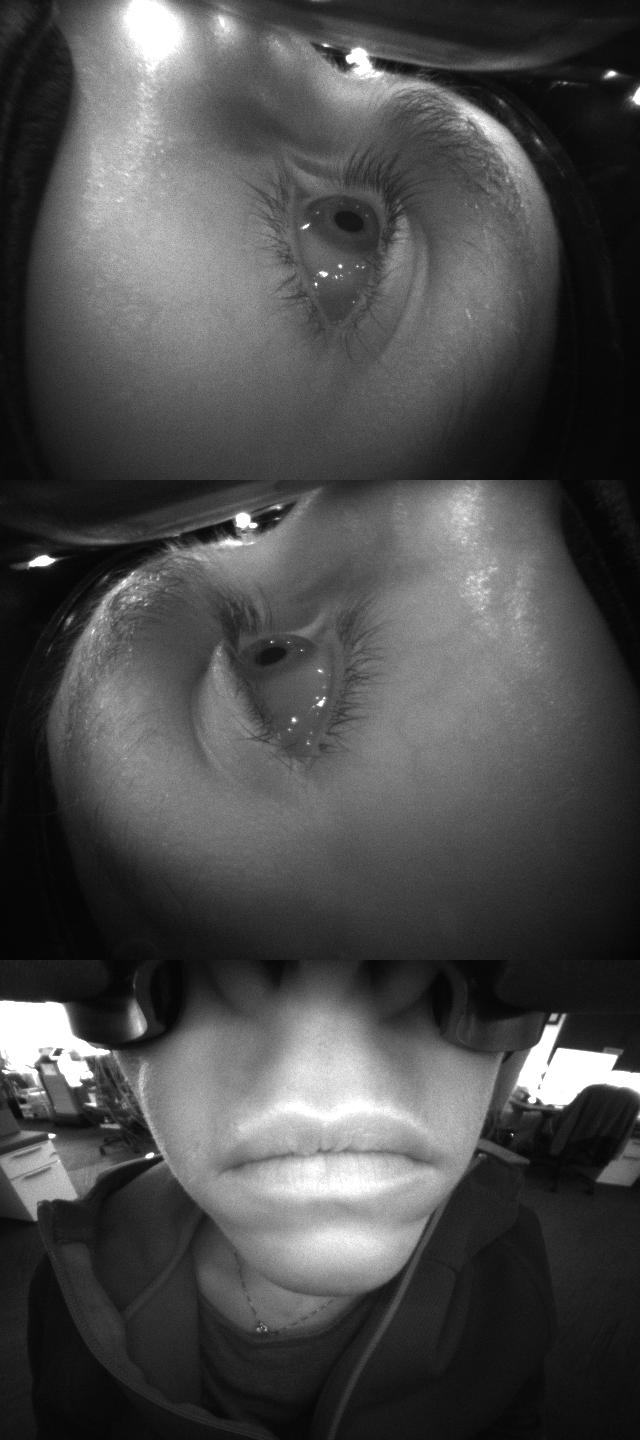} &
    \includegraphics[height=0.209\textwidth]{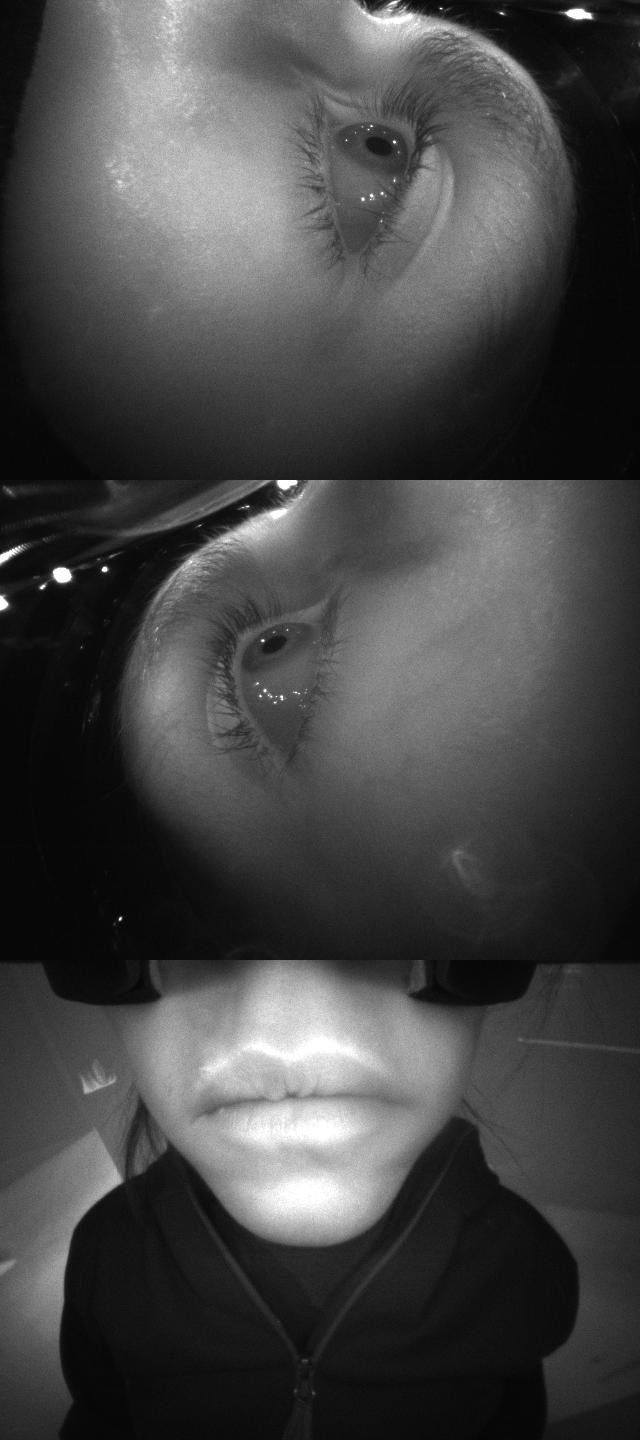} & 
    \includegraphics[height=0.209\textwidth]{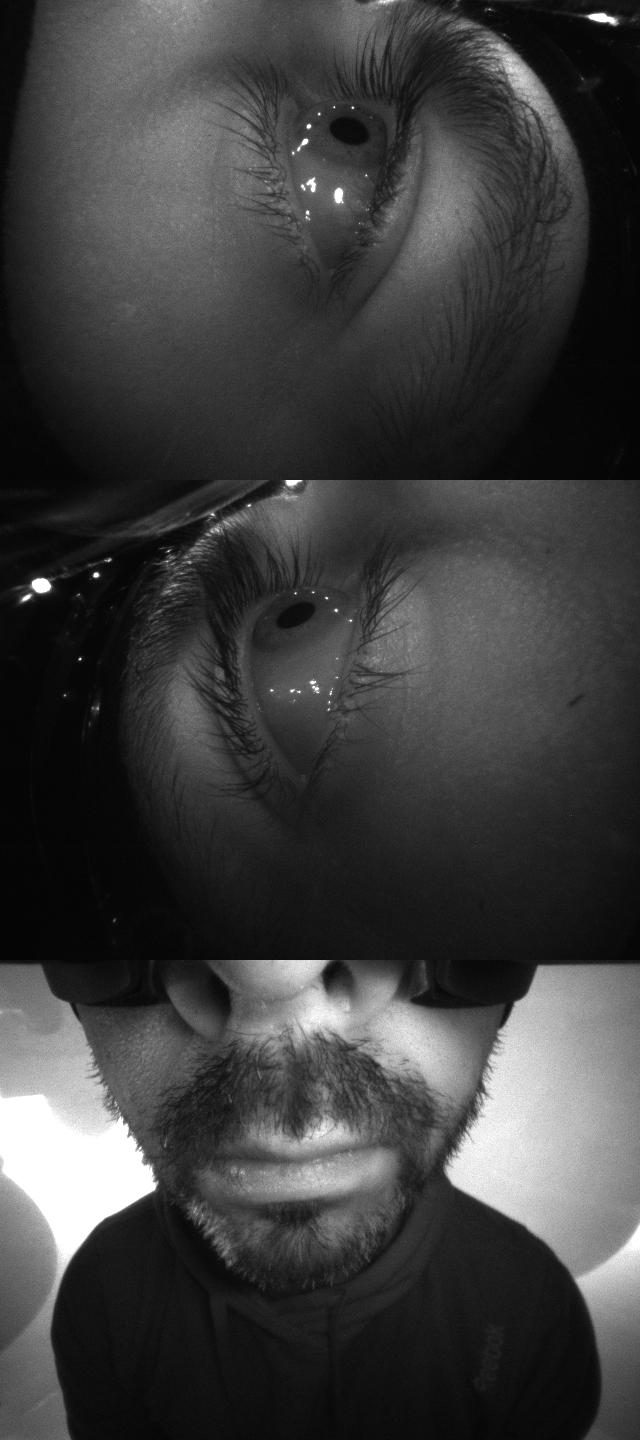} & 
    \includegraphics[height=0.209\textwidth]{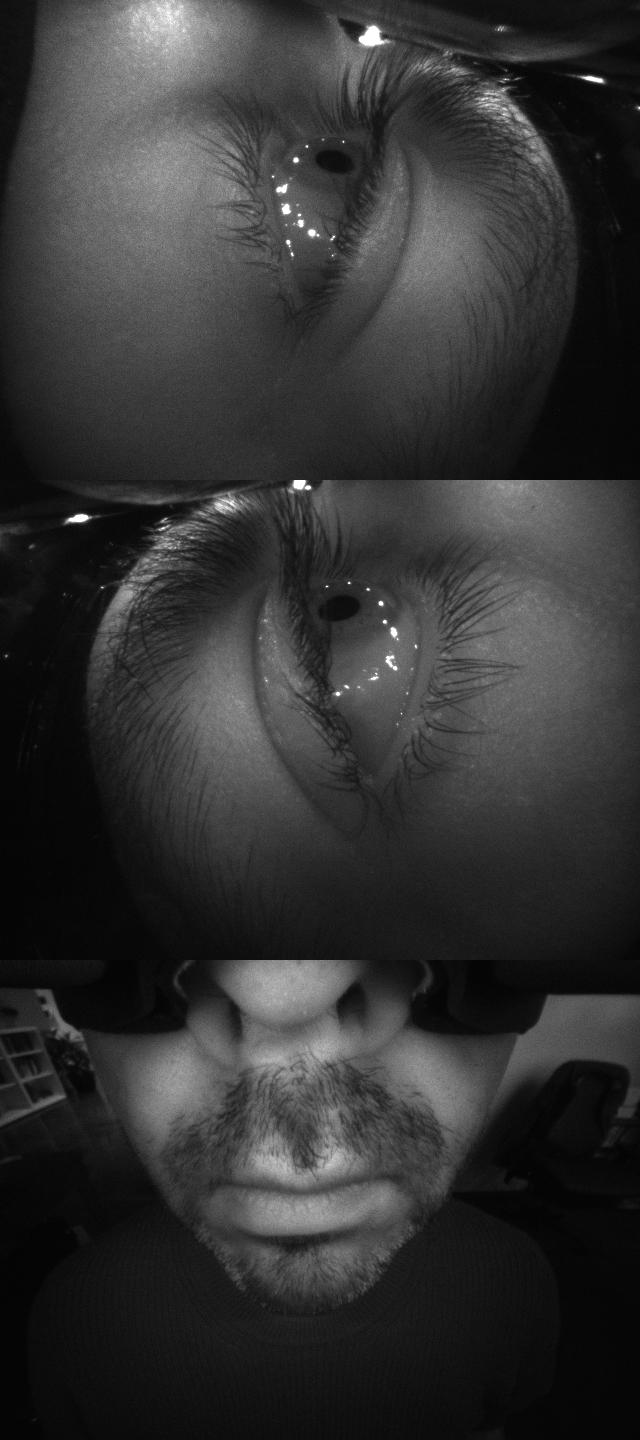} &
    \includegraphics[height=0.209\textwidth]{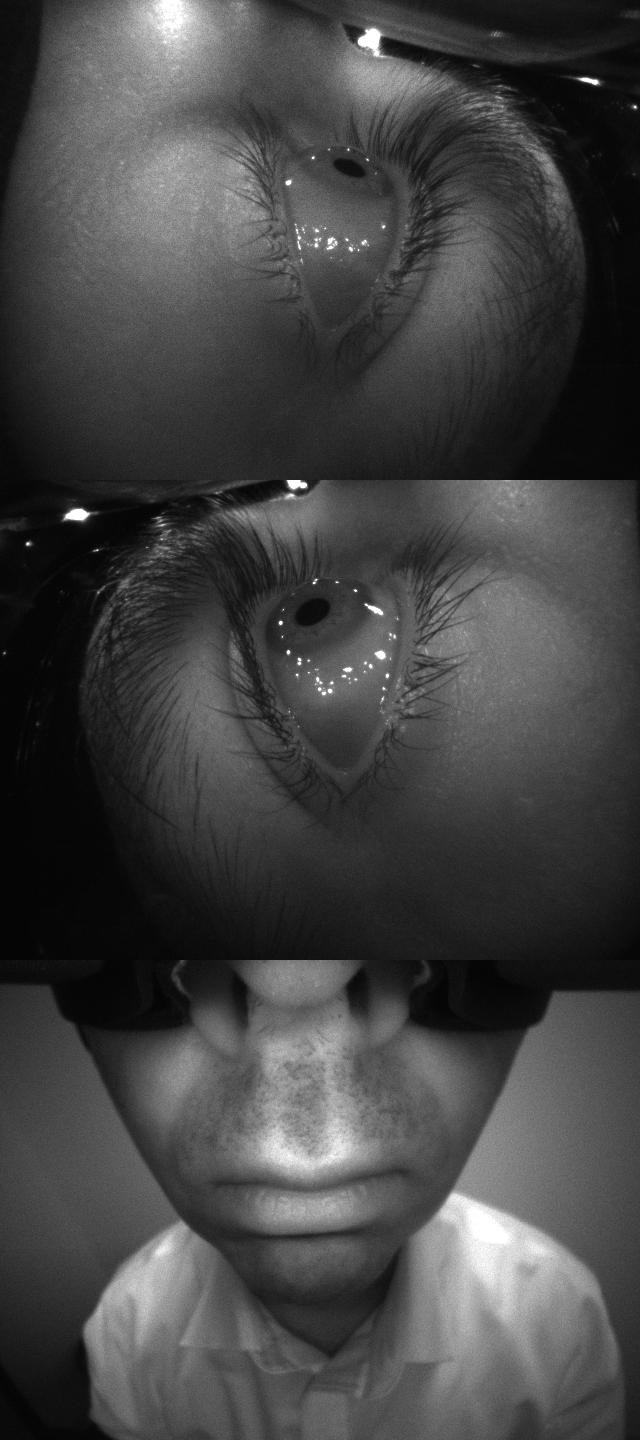} &
    \includegraphics[height=0.209\textwidth]{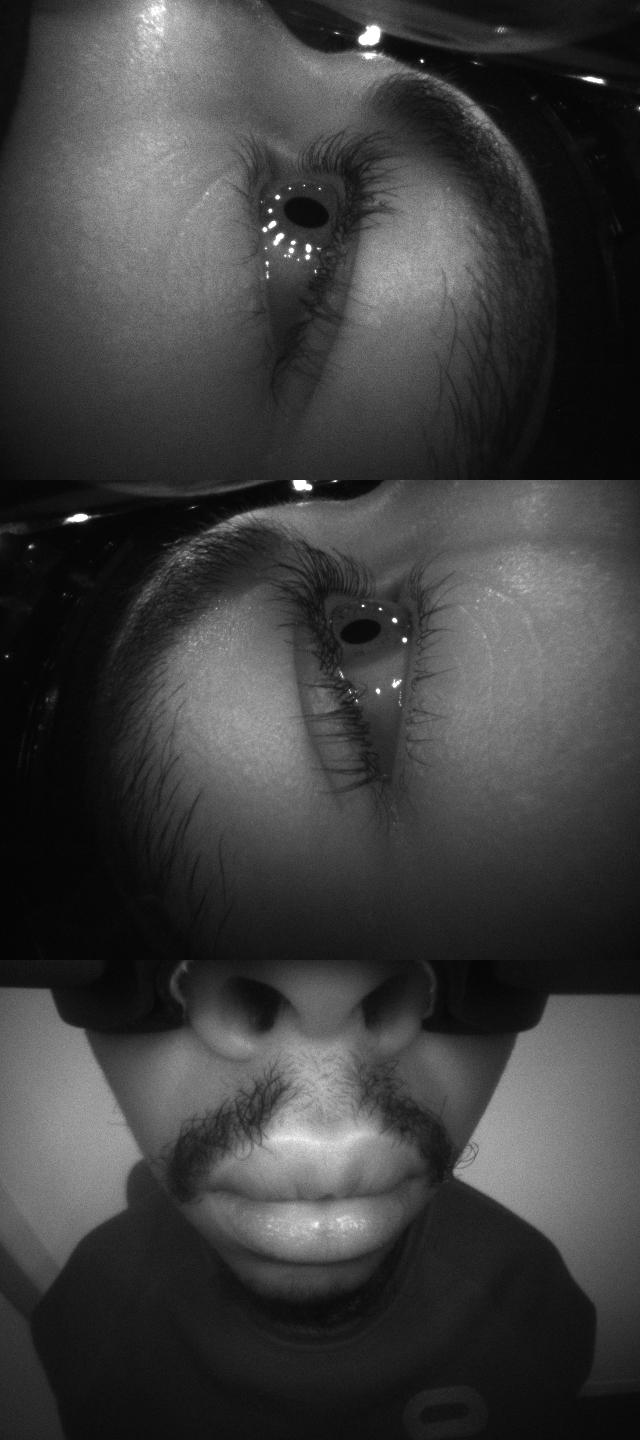} &
    \includegraphics[height=0.209\textwidth]{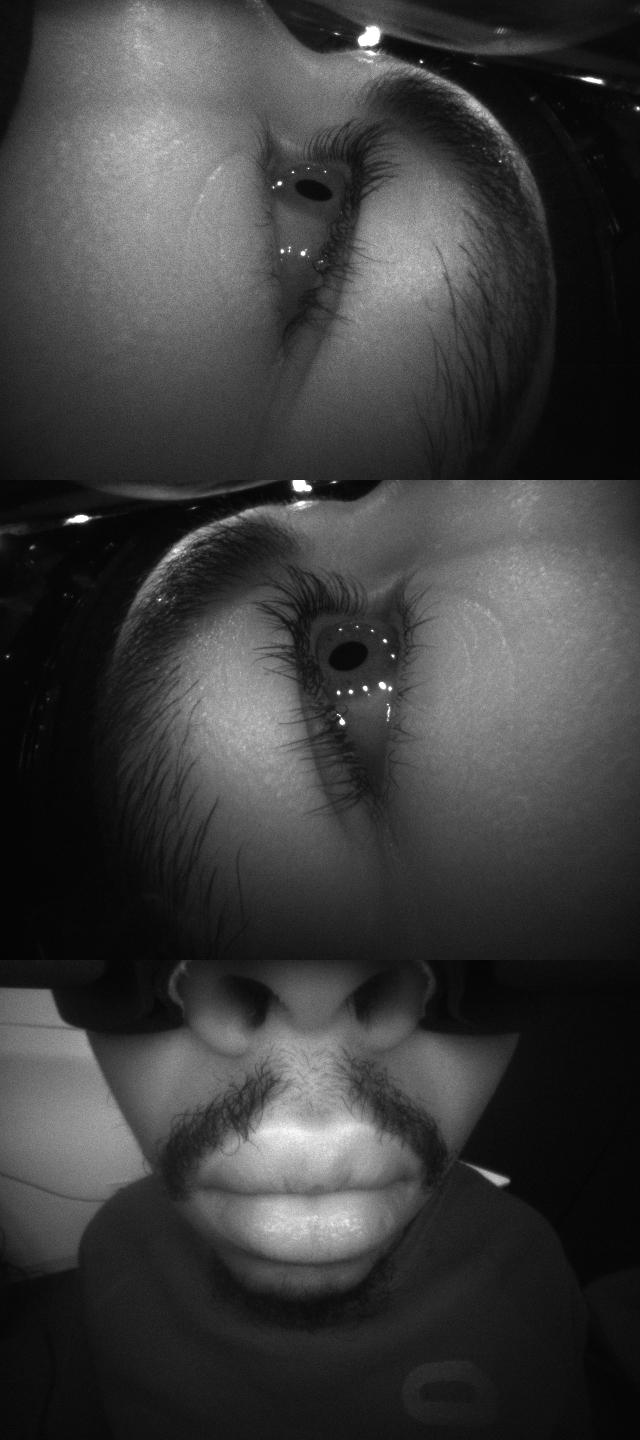} &
    \includegraphics[height=0.209\textwidth]{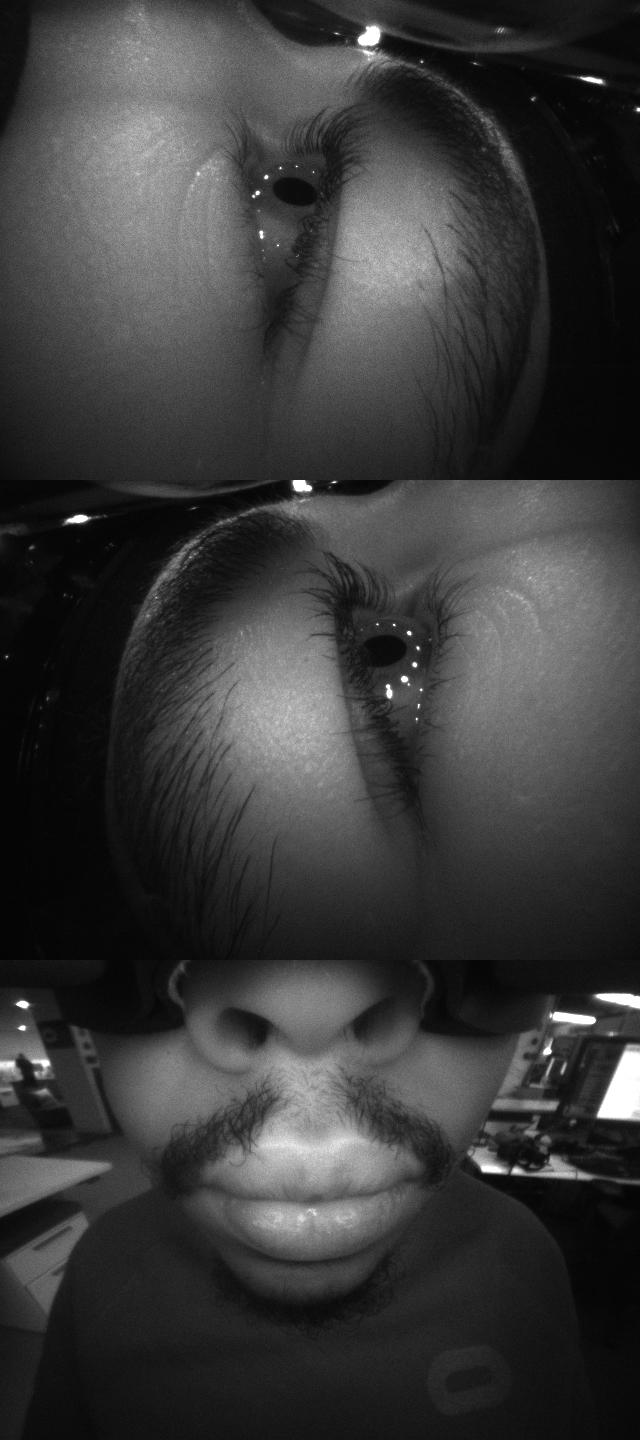} &
    \includegraphics[height=0.209\textwidth]{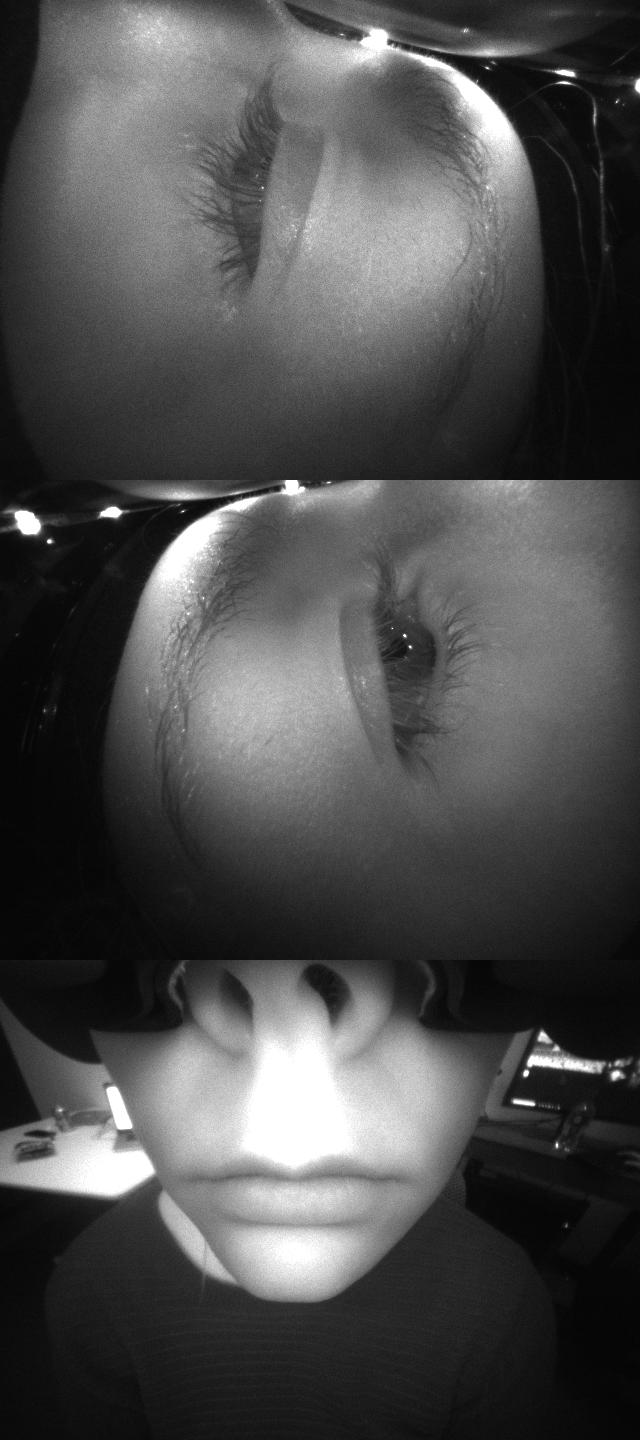} &
    \includegraphics[height=0.209\textwidth]{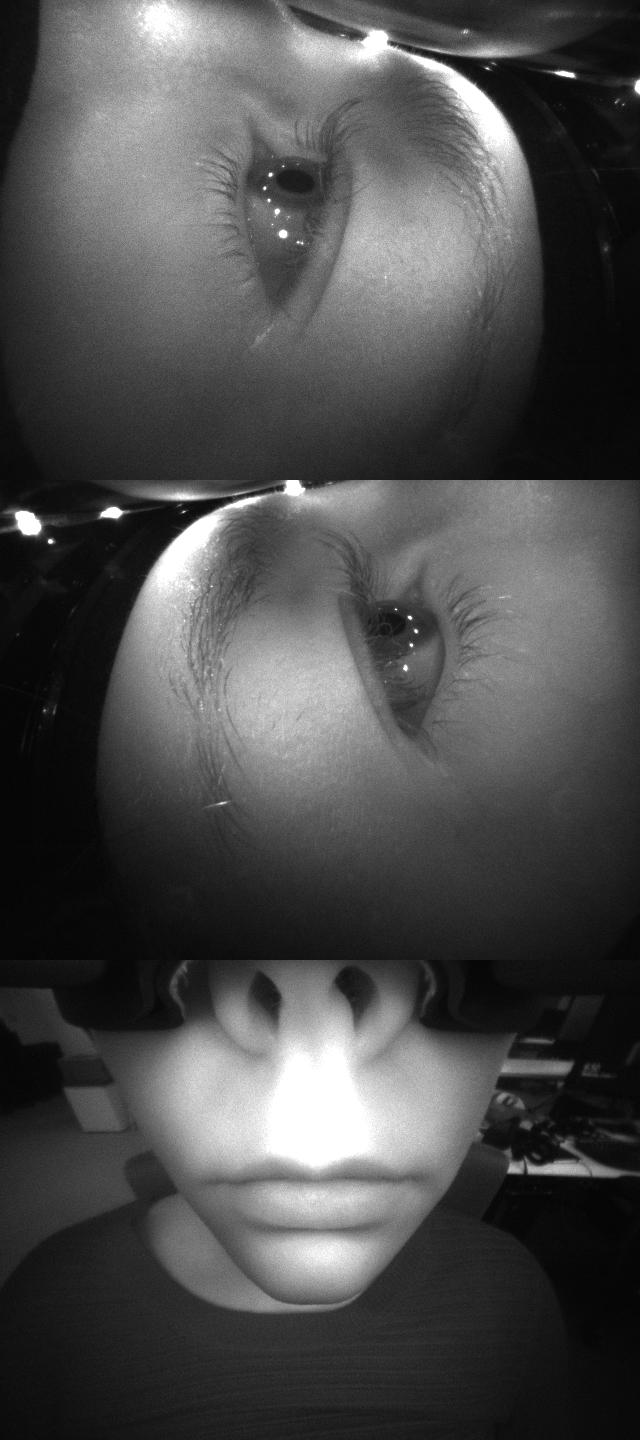}\\
    \end{tabu}
    \\\specialrule{.1em}{.05em}{.05em}
    \rotatebox[origin=c]{90}{{\scriptsize GT}} &
    \tabulinesep=\tabcolsep
    \begin{tabu} to 0.97\textwidth {X[c,m]X[c,m]X[c,m]X[c,m]}
    \includegraphics[height=0.2125\textwidth]{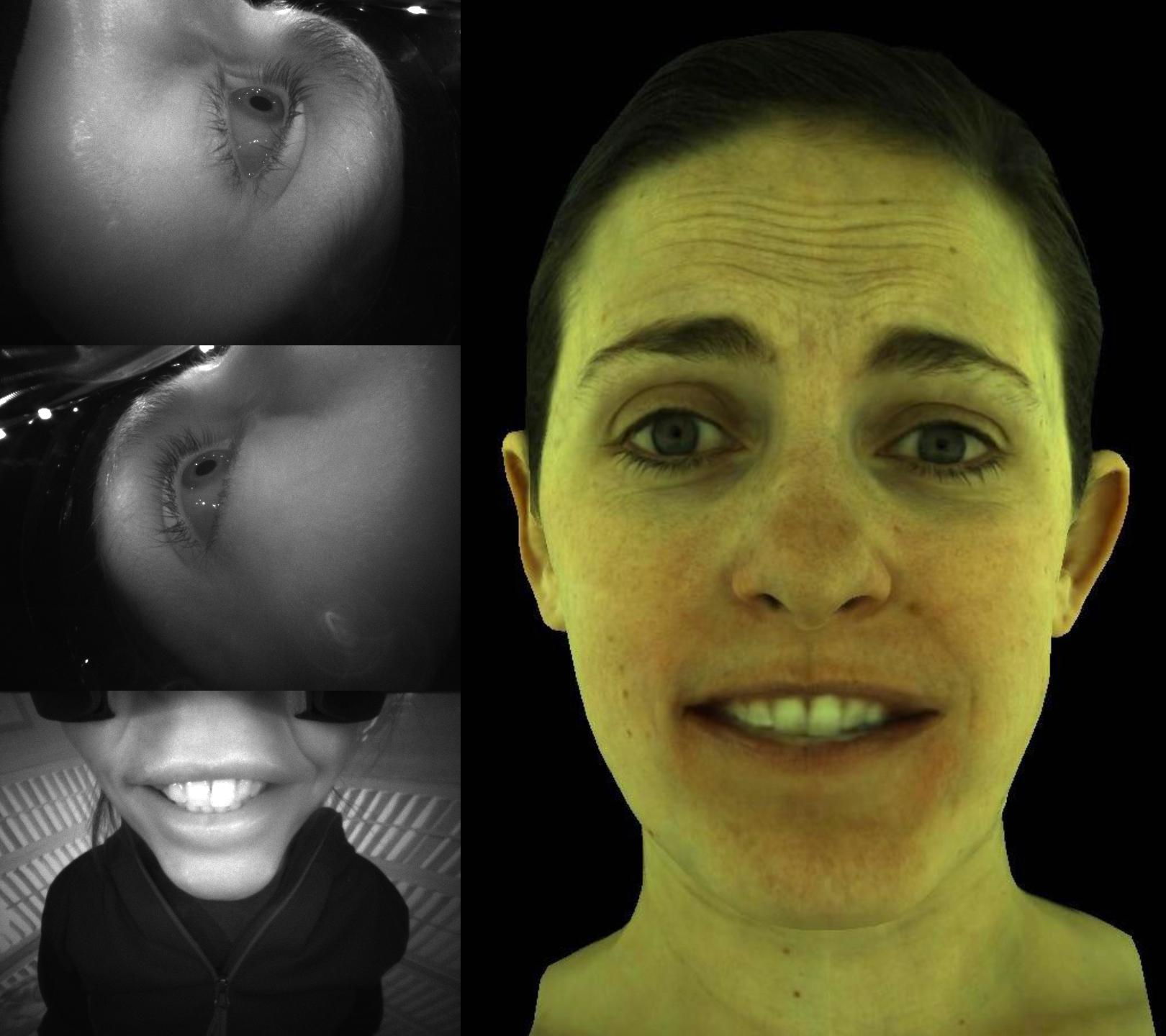} &
    \includegraphics[height=0.2125\textwidth]{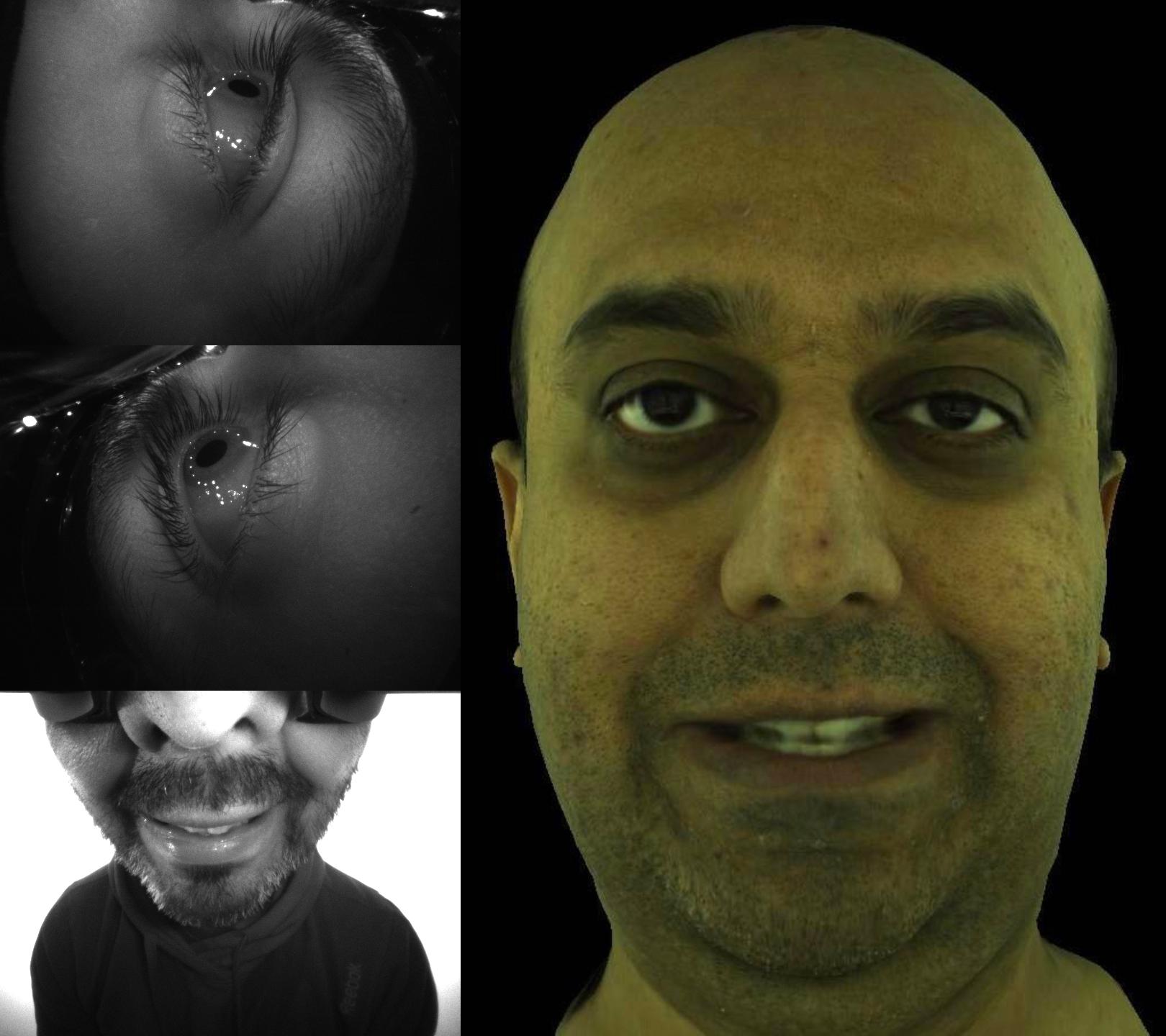} & 
    \includegraphics[height=0.2125\textwidth]{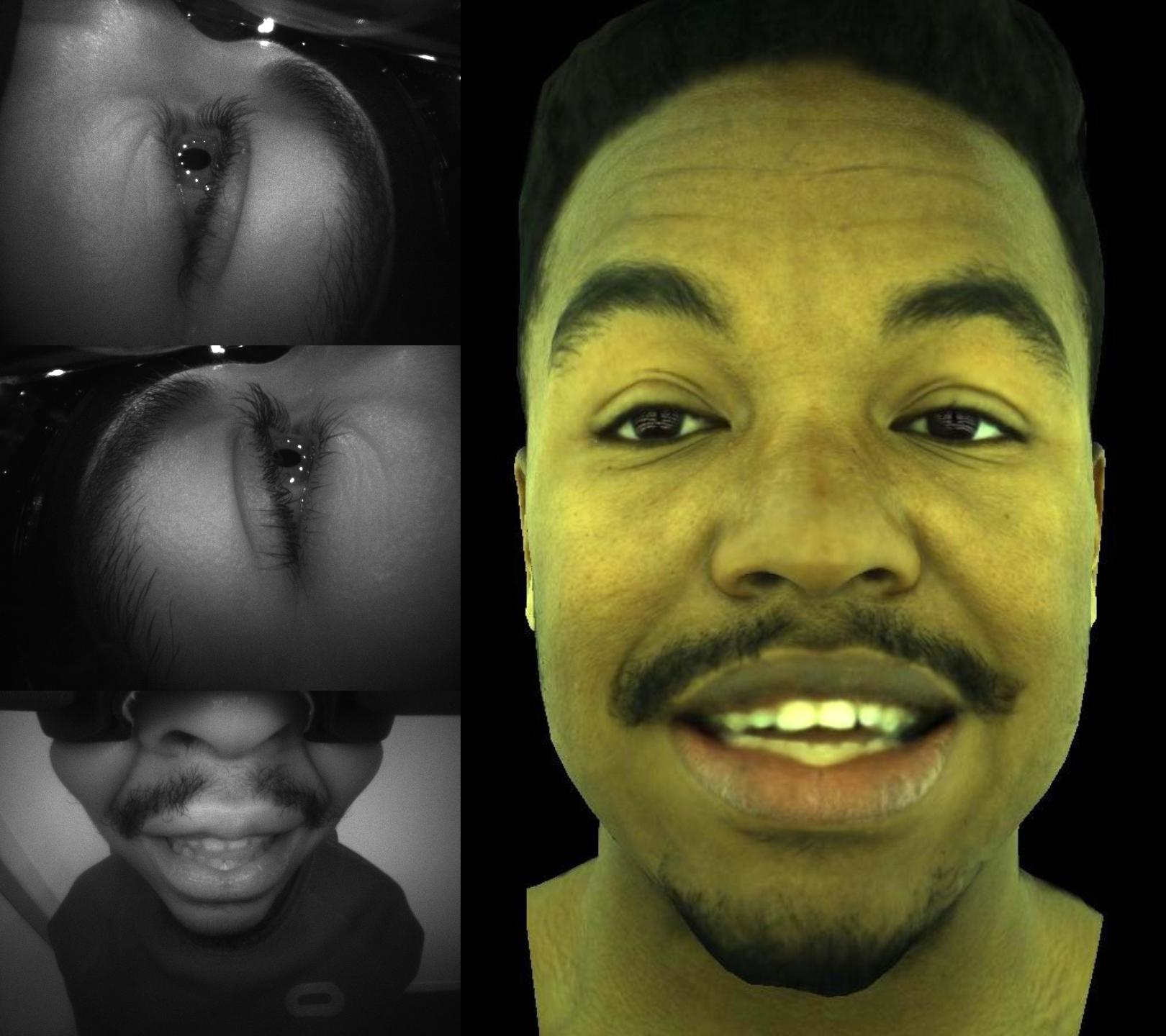} & 
    \includegraphics[height=0.2125\textwidth]{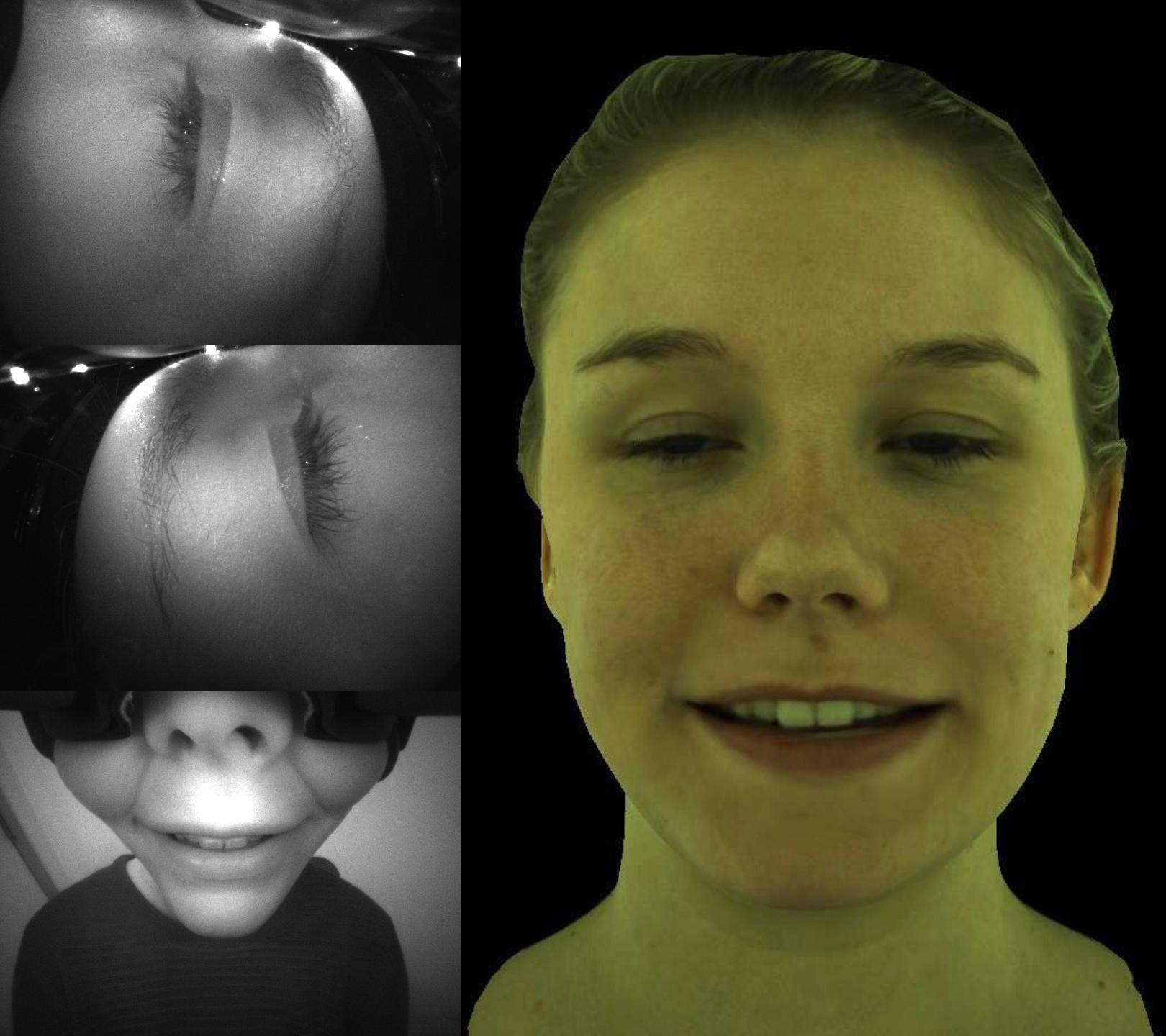}\\
    \end{tabu}
    \\\specialrule{.1em}{.05em}{.05em}
    \rotatebox[origin=c]{90}{{\scriptsize stat.}} &
    {\scriptsize
    \begin{tabular}{x{16.3mm}x{16.3mm}x{16.3mm}x{16.3mm}x{16.3mm}x{16.3mm}x{16.3mm}}
    \textbf{headset} & \textbf{person} & \textbf{room} & \textbf{capture} & \textbf{video} & \textbf{frame} & \textbf{image}\\\midrule
    3 & 4 & 6 & 14 & 1,691 & 203,729 & 1,833,561\\
    \end{tabular}
    }\\\specialrule{.1em}{.05em}{.05em}
\end{tabular}
\caption{\footnotesize Hardware and dataset examples. First row: capture dome~\cite{lombardi2018deep}, training headset, tracking headset, and sample images~\cite{wei2019vr}. Second row: head-mounted camera examples of left eye, right eye, and mouth from different capture runs. Third row: examples of head-mounted images and ground-truth correspondences. Last row: dataset statistics.}
\label{fig:dataset}
\end{figure}

\noindent
\textbf{Morphable 3D Facial Models:} Part-based  models have been widely used for modeling facial appearance because of their elasticity and the ability to handle occlusion. 
Cootes et al.~\cite{cootes2001active} introduced Active Appearance Models (AAM) for locating deformable objects such as faces. The facial shape and part-based appearance are iteratively matched to the image, using parameter displacement estimated from residual errors. 
The 3D morphable face model by Blanz and Vetter~\cite{blanz1999morphable} decomposes the face into four parts to augment expressiveness, despite the fact that PCA decomposition is still computed holistically on the whole face. 
Tena et al.~\cite{tena2011interactive} proposed region-based face models that use a collection of PCA sub-models with shared boundaries. Their method achieved semantically meaningful expression bases, while generalising better to unseen data compared to the holistic approach.
Neumman et al.~\cite{neumann2013sparse} extended sparse matrix decomposition theory to face mesh sequence processing, and a new way to ensure spatial locality.
Cao et al.~\cite{cao2018stabilized} proposed to learn part-based rigidity prior from existing facial capture data~\cite{cao2013facewarehouse}, which was used to impose rigid stability and avoid head pose jittery in real-time animation.
Recently, Ghafourzadeh et al.~\cite{ghafourzadeh2020part} presented local PCA-based model combined with anthropometric measurement to ensure expressiveness and intuitive user control.
Our approach also decomposes the face into part-based modules. However, instead of using linear or shallow features on the 3D mesh, our modules take place in latent spaces learned by deep neural networks. This enables capturing of complex non-linear effects, and producing facial animation with a new level of realism.

\noindent
\textbf{Deep Codec Avatars:} Human perception is particularly sensitive to detecting the realism of facial animation, e.g. the well-known Uncanny Valley Effect~\cite{seyama2007uncanny}. Traditional approaches such as morphable 3D models usually fail to pass the uncanny valley. Recent deep codec avatars have brought new hope to overcome this issue.
Lombardi et al.~\cite{lombardi2018deep} introduced Deep Appearance Models (DAM) that use deep Variational Auto-Encoders (VAE)~\cite{kingma2013auto} to jointly encode and decode facial geometry and view-dependent appearances into a latent code. The usage of deep VAE enabled capturing complex non-linear animation effects, while producing a smooth and compact latent representation. View-dependent texture enabled modeling view-dependent effects such as specularity, as well as allowing correcting from imperfect geometry estimation. Their approach enabled realistic facial animation without relying on expensive light-transport rendering.
Recently, Lombardi et al.~\cite{lombardi2019neural} extended their prior work to Neural Volumes (NV). They presented a dynamic, volumetric representation learned through encoder-decoder neural networks using a differentiable ray-marching algorithm. Their approach circumvents the difficulties in conventional mesh-based representation and does not require explicit reconstruction or tracking.
The work in \cite{lombardi2018deep} is an important cornerstone that our work is built upon. Our work differs in that our main focus is producing robust and accurate facial animations in realtime.

\noindent
\textbf{VR Telepresence:} The previous work that is most closely related to ours is Wei et al.~\cite{wei2019vr}. 
VR telepresence presents many unique challenges due to its novel hardware setup, e.g. unaccommodating camera views that can only see parts of the face, as well as the image domain gap from head-mounted cameras to realistic avatars. 
In Thies et al.~\cite{thies2016face2face} and Cao et al.~\cite{cao2014displaced}, a well-posed, unobstructed camera is used to provide image input for realtime reenactment. VR telepresence is different in that clear view of the full face is unavailable because the user has to wear the VR headset. 
Li et al.~\cite{li2015facial} presented an augmented VR headset that contains an outreaching arm holding an RGB-D camera.
Lombardi et al.~\cite{lombardi2018deep} used cameras mounted on a commodity VR headset, where telepresence was achieved by image-based rendering to create synthetic head-mounted images, and learning a common representation of real and synthetic images.
Wei et al.~\cite{wei2019vr} extended this idea by using CycleGAN~\cite{zhu2017unpaired} to further alleviate the image domain gap. A training headset with 9 cameras was used to establish correspondence, while a standard headset with 3 cameras was used to track the face.
Existing work in VR telepresence are based on holistic face modeling. In our work, we revive the classic module-based approach and combine it with codec avatars to achieve expressive and robust VR telepresence.

%% file: 3_dataset.tex
\section{Hardware \& Dataset}
In this section, we first describe our hardware setup. After that, we provide more details about the dataset that we use to train and evaluate our model.

Gathering high quality facial image data is the foundation of realistic facial animation. For this purpose, we use a large multi-camera dome that contains 40 cameras capturing images at a resolution of 2560$\times$1920 pixels. Cameras lie on the frontal hemisphere of the face at a distance of about 1 meter. The cameras are synchronized with a frame rate of 30fps. 200 LED lights are directed at the face to create uniform illumination.
To collect headset images, we used a two-headset design that has a training headset and a tracking headset. The training headset contains 9 cameras, which ensures establishing high-quality expression between head-mounted cameras and the avatar. The tracking headset contains a subset of 3 cameras, it is a consumer-friendly design with minimally intrusive cameras for real-time telepresence. The images are captured by IR cameras with a resolution of 640$\times$480 and frame rate of 30fps (down-sampled from 90fps).
We refer to \cite{lombardi2018deep} and \cite{wei2019vr} for more details about the camera dome and the VR headsets.

To obtain the high-fidelity facial avatar, we train a person-specific view-dependent VAE using 3D face meshes reconstructed and tracked from dome-captured data similar to \cite{lombardi2018deep}. The mesh contains 7306 vertices and a texture map of 1024$\times$1024. 
We then use the method in \cite{wei2019vr} to establish correspondences between training headset frames and avatar latent codes. We decode these corresponding latent codes into 3D face meshes. These serve as the ground truth outputs in our dataset. The 9-camera training headset is only used for obtaining accurate ground truth. Both training and evaluation of our model in later sections only uses the subset of 3 cameras on the tracking headset. Note that the method for establishing correspondences provides accurate ground truth output, but it is infeasible for real-time animation due to its expensive computational complexity.

We construct a dataset that covers common variations in practical usage scenarios: varying users, varying headsets and varying environments with different backgrounds and lighting conditions. Fig.~\ref{fig:dataset} shows some example data and statistics of the dataset. Specifically, our dataset contains four different users. We used three different headsets, and captured a total of 14 sessions (half an hour for each session) in three different indoor environments: a small room with weak light, an office environment in front of a desk under bright white light, and in front of a large display screen. In the last environment, we capture some sections while playing random high-frequency flashing patterns on the screen, to facilitate the evaluation under extreme lighting condition. Sessions of the same person may be captured on different dates that are months apart, resulting in potential appearance changes in the captured data. For example, for one person, we captured him with heavy beard in some sessions and the other sessions are captured after he shaved.

For each dome capture or headset capture, we recorded a predefined set of 73 unique facial expressions, recitation of 50 phonetically-balanced sentences, two range-of-motion sequences where the user is asked to move jaw or whole face randomly with maximum extend, and 5-10 minutes conversation. Finally, we split the dataset by assigning one full headset capture session of each person as the testing set. This means the testing set does not have overlap in terms of capture sessions. We use only sequences of sentence reading and conversation for testing as they reflect the usual facial behaviors of a user in social interactions. 

%% file: 4_method.tex
\section{Method}
The goal of our head-only VR telepresence system is to faithfully reconstruct full faces from images captured by the headset-mounted cameras in realtime. In this section, we will first describe the formulation of our model, followed by two important techniques that are important for successfully training the models and lastly implementation details.

\subsection{Model Formulation}
We denote the images captured by the headset-mounted cameras at each time instance as $\mathbf{X}$=$\{\mathbf{x}_k\ |\ k \in \{1,...,K\}\}$, with $K$ the total number of cameras and $\mathbf{x}_k$$\in$$\mathrm{I\!R}^{I}$ where $I$ is the number of pixels in each image. Note the time subscript $t$ is omitted to avoid notation clutter. We denote  $\mathbf{v}$$\in$$\mathrm{I\!R}^{3}$ as the direction from which the avatar is to be viewed in VR.
Given $\mathbf{X}$, the telepresence system needs to compute the view-dependent output $\mathbf{y}_{\mathbf{v}}$ which contains the facial geometry $\mathbf{y}^g$$\in$$\mathrm{I\!R}^{3G}$ of 3D vertex positions and the facial texture $\mathbf{y}_{\mathbf{v}}^t$$\in$$\mathrm{I\!R}^{3T}$ corresponding to view direction $\mathbf{v}$. $G$ and $T$ are the number of vertices on the facial mesh and number of pixels on the texture map respectively. 

The holistic Codec Avatar (CA)~\cite{wei2019vr,lombardi2018deep} is formulated as a view-dependent encoder-decoder framework, i.e.
\begin{align}
    \mathbf{y}_{\mathbf{v}}=\mathcal{D}\left(\mathbf{c},\mathbf{v}\right), \;
    \mathbf{c}=\mathcal{E}\left(\mathbf{X}\right)
\end{align}
where headset cameras' images $\mathbf{X}$ are feed into an image encoder $\mathcal{E}$ to produce the expression code of the whole face, and a decoder $\mathcal{D}$ produces the view-dependent 3D face. 
$\mathcal{D}$ is trained on dome-captured data to ensure animation quality, and $\mathcal{E}$ is trained using the correspondences between $\mathbf{X}$ and $\mathbf{c}$ that are established using the method in~\cite{wei2019vr}. Because $\mathcal{D}$ is a neural network trained with a limited set of facial expressions, CA has limited expressiveness in out-of-sample expressions due to the long tail distribution nature of human facial expressions. In this work we propose Modular Codec Avatar (MCA), where the 3D face modules are estimated by an image encoder followed by a synthesizer from each headset camera view, and blended together to form the final face. MCA can be formulated as:
\begin{align}
    \mathbf{y}_{\mathbf{v}}=\sum_{k=1}^{K}\mathbf{w}_k\odot \mathcal{D}_k\left(\mathbf{c}_k^{\mathrm{full}},\mathbf{v}\right)
\end{align}
\begin{align}
 \left[\mathbf{c}_k^{\mathrm{full}}, \mathbf{w}_k\right]=\mathcal{S}_k\left(\mathbf{c}_k^{\mathrm{part}}\right), \; \; \; \mathbf{c}_k^{\mathrm{part}}=\mathcal{E}_k\left(\mathbf{x}_k\right)
\end{align}
where each camera view $\mathbf{x}_k$ is processed separately. This computation consists of three steps. 
Firstly, a modular image encoder $\mathcal{E}_k$ estimates the modular expression code $\mathbf{c}_k^{\mathrm{part}}$ which only models the facial part visible in the $k$th camera view, e.g., left eye. 
Subsequently, a modular synthesizer $\mathcal{S}_k$ estimates a latent code for the full-face denoted as $\mathbf{c}_k^{\mathrm{full}}$, based only on the information from $\mathbf{c}_k^{\mathrm{part}}$. The synthesizer also estimates blending weights $\mathbf{w}_k$. 
Lastly, we aggregate the results from all $K$ modules to form the final face: decode the 3D modular faces using $\mathcal{D}_k$, and blend them together using the adaptive weights. $\odot$ represents the element-wise multiplication.
In this way, MCA learns part-wise expressions inside each face module, while keeping full flexibility of assembling different modules together.

The objective function for training the MCA model consists of three loss terms for the reconstruction and intermediate latent codes:
\begin{align}
    \mathcal{L}_{\mathrm{MCA}}=\left\|\mathbf{y}_{\mathbf{v}_0}-\hat{\mathbf{y}}_{\mathbf{v}_0}\right\|_2 + \lambda_1 \sum_{k=1}^{K} \left\|\mathbf{c}_k^{\mathrm{full}}-\hat{\mathbf{c}}\right\|_2 + \lambda_2 \sum_{k=1}^{K} \left\|\mathbf{c}_k^{\mathrm{part}}-\hat{\mathbf{c}}_k^{\mathrm{part}}\right\|_2
\end{align}
where $\hat{\mathbf{y}}_{\mathbf{v}_0}$, $\hat{\mathbf{c}}$, and $\hat{\mathbf{c}}_k^{\mathrm{part}}$ denote different supervision signals. The first term measures the reconstruction error in facial geometry and texture. We set $\hat{\mathbf{y}}_{\mathbf{v}_0}$=$\mathcal{D}(\hat{\mathbf{c}},\mathbf{v}_0)$ as the facial geometry and texture decoded from the supervision code $\hat{\mathbf{c}}$ from frontal view direction $\mathbf{v}_0$. Note the supervision $\hat{\mathbf{c}}$ is estimated from the correspondence stage using the method in~\cite{wei2019vr}. We impose equal weights for reconstruction error in geometry and texture. For the texture, we average pool the reconstruction errors of the pixels corresponding to the same closest mesh vertex instead of averaging over all pixels on the texture map. This vertex-based pooling ensures that texture loss has a geometrically uniform impact.
The second term encourages each module to produce independent estimation of the correct full-face and the last term directs each encoder to produce correct modular expression code. Section~\ref{sec:examplar} describes in detail how we generate the supervision $\hat{\mathbf{c}}_k^{\mathrm{part}}$.

\subsection{Exemplar-based Latent Alignment}
\label{sec:examplar}
The two latent codes $\mathbf{c}_k^{\mathrm{part}}$ and $\mathbf{c}_k^{\mathrm{full}}$ have different purposes. $\mathbf{c}_k^{\mathrm{part}}$ represents information only within its responsible modular region, while $\mathbf{c}_k^{\mathrm{full}}$ further synthesizes a full-face based on the single-module expression information. 
It is important to have $\mathbf{c}_k^{\mathrm{part}}$, because otherwise the modules will only collectively try to recover the same full-face code through $\mathbf{c}_k^{\mathrm{full}}$, which essentially degrades to the holistic CA. 
The key to ensure the effectiveness of $\mathbf{c}_k^{\mathrm{part}}$ is through crafting proper supervision signal $\hat{\mathbf{c}}_k^{\mathrm{part}}$. The main challenge is $\hat{\mathbf{c}}_k^{\mathrm{part}}$ resides in an inexplicitly defined latent space. We address this problem with exemplar-based latent alignment.

To obtain $\hat{\mathbf{c}}_k^{\mathrm{part}}$, we first train a separate VAE for each module from dome-captured data. It has a similar architecture as CA, but only uses the region within the module by applying a modular mask on both the VAE input and output.
We denote the masked modular VAE decoder as $\mathcal{D}_k^{\mathrm{mask}}$, and the set of codes corresponding to dome-captured modular faces as $\mathbf{C}_k^{\mathrm{mask}}$.
The main challenge to obtain $\hat{\mathbf{c}}_k^{\mathrm{part}}$ is the domain gap between dome-captured and headset-captured data. Directly applying the trained modular VAE on masked ground truth often result in spurious code vectors, i.e., the dome code and headset code corresponding to similar expression content do not match. This is caused by lighting and appearance differences between the two drastically different capture setup. Moreover, the domain gap also exist between different headset capture runs.
To overcome this mismatch, we use exemplar-based latent alignment that replace the headset-captured codes produced by the decoder by their nearest dome-captured exemplar code. This effectively calibrates $\hat{\mathbf{c}}_k^{\mathrm{part}}$ from different capture runs to a consistent base, i.e.
\begin{align}
    \hat{\mathbf{c}}_k^{\mathrm{part}}=\underset{\mathbf{c}\in \mathbf{C}_k^{\mathrm{mask}}}{\mathrm{argmin}}\left\|\mathcal{D}_k^{\mathrm{mask}}\left(\mathbf{c},\mathbf{v}_0\right)-\hat{\mathbf{y}}_{\mathbf{v}_0,k}^{\mathrm{mask}}\right\|_2 
\end{align}
where $\hat{\mathbf{y}}_{\mathbf{v}_0,k}^{\mathrm{mask}}$ is the modular masked $\hat{\mathbf{y}}_{\mathbf{v}_0}$. This differs from pure image-based synthesis in that result must come from a set of known dome-captured exemplars $\mathbf{C}_k^{\mathrm{mask}}$. The resulting $\hat{\mathbf{c}}_k^{\mathrm{part}}$ is then used in Eq.(4) to train MCA.

\subsection{Modulated Adaptive Blending}
The blending weights $\mathbf{w}_k$ can be fully learned from data. However, we find automatically learned blending weights lead to animation artifacts. This is because module correlation exists in training data, e.g. left and right eyes often open and close together. Therefore, the dominant module interchanges between left and right eyes across nearby pixels, which results in jigsaw-like artifacts in the eyeball region. To overcome this issue and promote spatial coherence in the modular blending weights, we add a multiplicative and additive modulation signals to the adaptively learned blending weights. This ensures blending weight is constant near the module centroid, and the importance of adaptively learned weights gradually increases, i.e.
\begin{align}
    \mathbf{w}_k=\frac{1}{\mathbf{w}}\odot\left(\mathbf{w}_{k}^{\mathcal{S}}\odot e^{-\frac{\mathrm{max}\{\left\|\mathbf{u}-\overline{\mathbf{u}}_{k}\right\|^2-a_k,0\}}{\sigma^2}}+b_k \mathds{1}\left\{\left\|\mathbf{u}-\overline{\mathbf{u}}_{k}\right\|^2\leq a_k\right\}\right)
\end{align}
where $\mathbf{w}_{k}^{\mathcal{S}}$ denotes the adaptive blending weights produced by the synthesizer $\mathcal{S}_k$, $\mathbf{u}$ denotes the 2D texture map coordinates corresponding to each vertex, $\overline{\mathbf{u}}_{k}$, $a_k$, and $b_k$ are constants denoting the module's centroid, area in the texture map, and constant amplitude within its area. $\mathbf{w}$ is computed vertex-wise to normalize the blending weights across all modules.

\subsection{Implementation Details}
We use $K$=3 modules following our hardware design, with three head-mounted cameras capturing left eye, right eye, and mouth.
Each $\mathbf{x}_k$ is resized to a dimension of 256$\times$256.
We use a latent dimension of 256 for both $\hat{\mathbf{c}}_k^{\mathrm{part}}$ and $\hat{\mathbf{c}}$.
Similar neural network architectures to \cite{wei2019vr}  are used for our encoder $\mathcal{E}_k$ and decoder $\mathcal{D}_k$. For the decoder, we reduce the feature dimensions to remedy computation cost due to decoding $K$ times. We also set all $\mathcal{D}_k$ to share the same weights to save the capacity requirement for storage and transmission of the model.
The synthesizer $\mathcal{S}_k$ consists of three temporal convolution layers~\cite{bai2018empirical} (TCN), with connections to a small temporal receptive field of 4 previous frames.
The blending weights $\mathbf{w}_k^{\mathcal{S}}$ is predicted through transposed convolution layers with a sigmoid function at the end to fit the texture map resolution. The texture weights were then reordered to produce geometry weights using the correspondence between vertices and texture map coordinates.
We also add cross-module skip connections that concatenates the last layer features in $\mathcal{E}_k$ to allow the model to exploit correlations between images from different headset-mounted cameras.

We train the model with the Adam optimizer with both $\lambda_1$ and $\lambda_2$ set to 1. We augment the headset images by randomly cropping, zooming, and rotating the images. We also add random gaussian noise to the modular code $\hat{\mathbf{c}}_k^{\mathrm{part}}$ with diagonal covariance determined by the distance to the closest neighbour to prevent overfitting. The training is completed using four Tesla V100 GPUs. During test time, the telepresence system using MCA takes in average 21.6ms to produce one frame in VR, achieving real-time photo-realistic facial animation.

%% file: 5_experiment.tex
\section{Experiments}
The first experiment illustrates the advantage of MCA over CA modeling untrained facial expressions. We then provide detailed evaluation results of the full VR telepresence via MCA comparing to CA. Our experiment evaluates the performance from extensive perspectives, including expression accuracy and perceptive quality. We also provide detailed ablation study and discuss failure cases. Finally, we show two extensions of our method which may be useful under certain usage scenarios.

\subsection{Facial Expression Modeling: Modular (MCA) vs. Holistic (CA)}
\begin{figure}[!t]
    \centering
    \setlength{\tabcolsep}{0pt}
    \begin{tabular}{cccc}
    \includegraphics[width=0.25\textwidth]{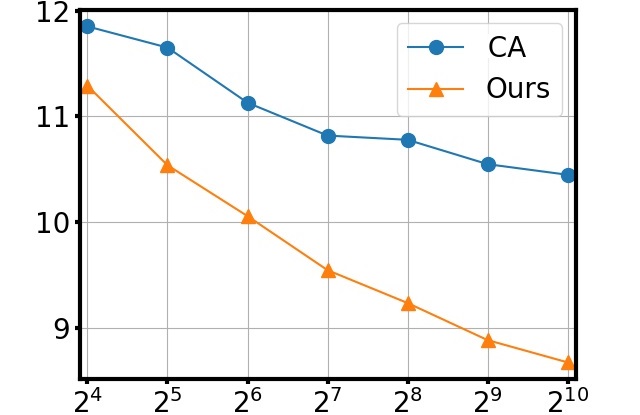} & 
    \includegraphics[width=0.25\textwidth]{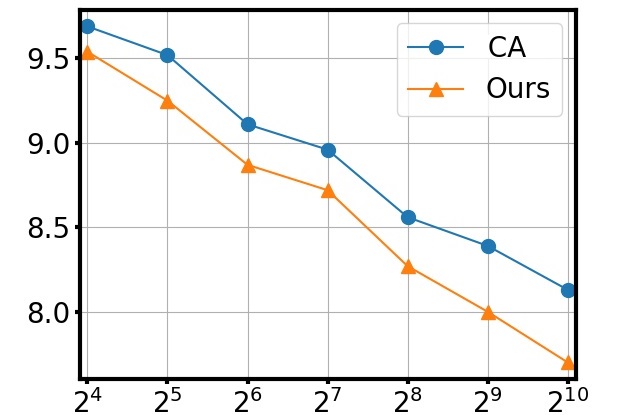} &
    \includegraphics[width=0.25\textwidth]{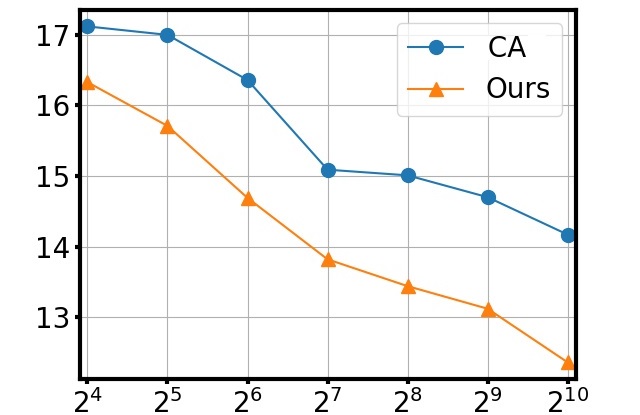} & 
    \includegraphics[width=0.25\textwidth]{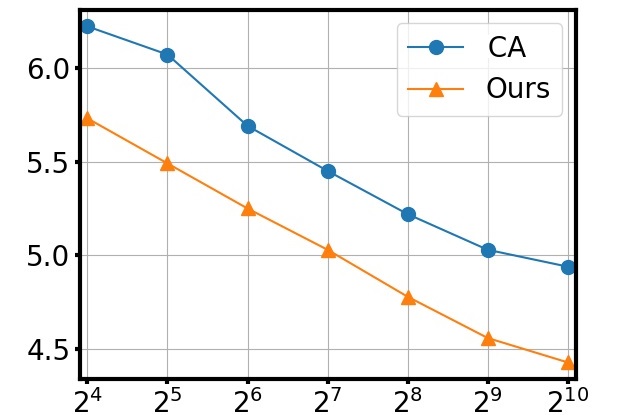}\\
    {\scriptsize person1} & {\scriptsize person2} & {\scriptsize person3} & {\scriptsize person4}\\ 
    \end{tabular}
    \caption{\footnotesize Comparing holistic versus modular by isolating and evaluating the importance expressiveness. X-axis shows different capacities of the face expressions by the number of clusters. Y-axis shows the RMSE photometric error. Note that CA uses proportionally $K$ times more clusters than each module of MCA for fair comparison.}
    \label{fig:expressive}
\end{figure}

In the first experiment, we show the advantage of modeling modular expressions in real world VR telepresence scenarios.
Towards this goal, we train a holistic VAE~\cite{wei2019vr,lombardi2018deep} as well as modular VAEs on dome-captured data. 
For both approach, we apply agglomerative clustering on the resulting latent codes of the training set data with varying number of clusters to represent the corresponding model's capacity of representing facial expressions. Recall that for headset-captured data, we have their estimated \textit{ground truth} 3D face through the correspondence stage using the method in~\cite{wei2019vr}. We then use these 3D faces to retrieve the closest cluster centers by matching the pixel values within each modular region of the rendered frontal faces, and report the root mean squared error (RMSE) of the pixel values in Fig.~\ref{fig:expressive}. For fair comparison, we use $K$=3 times more clusters for the CA baseline. Fig.~\ref{fig:expressive} shows the result.

It can be seen from Fig.~\ref{fig:expressive} that MCA consistently produces lower matching errors than CA throughout all subjects and model capacity levels. Intuitively this implies that MCA generalizes better to untrained expressions. The gap between CA and MCA increases as the number of clusters increases, indicating MCA's advantage increases as the face modeling becomes more fine-grained. It is then clear that by modeling modular expressions in the MCA, we can more truthfully interpolate/extrapolate facial expressions that are not contained in the dome-captured data, achieving better expressiveness in telepresence.      

\subsection{Full VR Telepresence}
We evaluate the performance of the full VR telepresence system. We feed headset images as input to the model, and evaluate the 3D face output against the ground-truth. Six metrics are used for comprehensive evaluation and the results are reported in Table~\ref{tab:quantitative}. 

\noindent
\textbf{Expression Accuracy} The most important metrics are MAE (i.e. Mean Absolute Error) and RMSE on pixels in the rendered frontal view face, as they directly and metrically measure the accuracy of VR telepresence. To further decompose the error into geometry and texture components, we compute RMSE on vertex position (Geo.) and texture map (Tex.) respectively as well. MCA \textit {consistently outperform} CA on all these metrics on almost all identities' test data. These results suggest that MCA can more truthfully recover the 3D face given only the partial views of the face from headset-camera images than CA, leading to more expressive and robust telepresense.

\noindent
\textbf{Improvement Consistency} As these metric values are computed from average of all frames, we need to verify that the improvement of MCA is not due to large performance differences on only a small set of frames. To do that, we compute the percentage of frames for which MCA produces more accurate results than CA and vice versa. This is reported as \%-better in Table~\ref{tab:quantitative}. From the results, it is clear that MCA \textit{consistently outperform}  CA across frames. For example, for person3, on 99.7\% frames MCA produces more accurate results than CA. 

\noindent
\textbf{Perceptive Quality} We also want to verify that such accuracy improvements align with human perception, i.e. whether a user may actually feel the improvement. Quantitatively, we compute structural similarity index on the grey-level frontal rendering (SSIM)~\cite{wangzhou2004} which is a perception-based metric that considers image degradation as perceived change in structural information, while also incorporating important perceptual phenomena~\cite{SSIM}. These results are reported in the last column in Table~\ref{tab:quantitative}: MCA outperforms CA on three persons while is on-par on the last one. This hints that the accuracy improvements of MCA over CA align with perception improvement. Fig.~\ref{fig:qualitative} shows a qualitative comparison between MCA and CA. It can be seen that MCA is better at handling subtle combined expressions, e.g. mouth open while eyes half-open, showing teeth while eyes shut, and mouth stretched while looking down. Renderings from different viewpoint by varying $\mathbf{v}$ are shown in Fig.~\ref{fig:viewpoint}. It can be seen that MCA produces consistent results from different viewing directions.    

\noindent
\textbf{Albation Study} Table~\ref{tab:ablation} shows an ablation study of MCA. Ablation factors range from the usage of synthesized blending weights instead of equal weights (blend), training the network end-to-end (end2end), using soft latent part vectors with gaussian noise instead of hard categorical classes (soft-ex.), expanding the dimension of latent codes (dimen.), using skip-module connections so all images are visible to the modular encoder (skip-mod.), and using temporal convolution on the synthesizer (tconv.). It can be seen the former three techniques improve performance significantly, while extra connections such as skip and temporal convolution lead to further improvements. We refer to supplemental material for more details.

\noindent
\textbf{Failure Cases} Fig.~\ref{fig:failure} shows example failure cases. Strong background flash is a challenging problem even for MCA, leading to inaccurate output (left of Fig.~\ref{fig:failure}). Although MCA can produce more expressive 3D faces, extreme asymmetric expressions like one pupil in the center while the other roll all the way to the corner as in the right of Fig.~\ref{fig:failure} still remains challenging to be faithfully reconstructed.

\begin{table}[!t]
    \centering
    \setlength{\tabcolsep}{4.2pt}
    {\scriptsize
    \begin{tabular}{c?{.1em}c|c?{.1em}c|c?{.1em}c|c?{.1em}c|c?{.1em}c|c?{.1em}c|c}
    \toprule
    ~ & \multicolumn{2}{c?{.1em}}{\textbf{MAE$\downarrow$}} & \multicolumn{2}{c?{.1em}}{\textbf{RMSE$\downarrow$}} &   \multicolumn{2}{c?{.1em}}{\textbf{Geo.$\downarrow$}} & \multicolumn{2}{c?{.1em}}{\textbf{Tex.$\downarrow$}} &
    \multicolumn{2}{c?{.1em}}{\textbf{\%-better$\uparrow$}} & \multicolumn{2}{c}{\textbf{SSIM$\uparrow$}} \\\specialrule{.1em}{.05em}{.05em}
    method & CA & Ours & CA & Ours & CA & Ours & CA & Ours & CA & Ours & CA & Ours \\\hline
    person1  & 8.82 & \textbf{8.69} & 7.67 & \textbf{7.47} & 1.26 & \textbf{1.14} & 3.40 & \textbf{3.02} & 36.3 & \textbf{63.7} & 0.954 & \textbf{0.957} \\
    
    person2 & 4.44 & \textbf{4.26} & 4.00 & \textbf{3.84} & 1.82 & \textbf{1.46} & 2.05 & \textbf{2.04}  & 27.3 & \textbf{72.7} & 0.949 & \textbf{0.951} \\
    
    person3  & 9.09 & \textbf{6.97} & 8.36 & \textbf{6.66} & 1.14 & \textbf{0.84} & 4.58 & \textbf{3.43} & 0.3 & \textbf{99.7} & 0.933 & \textbf{0.942}\\
    
    person4 & 3.33 & \textbf{3.21} & 3.08 & \textbf{3.04} & \textbf{0.54} & 0.64 & 0.86 & \textbf{0.85} & 41.1 & \textbf{58.9} & 0.984 & 0.984 \\\hline
    
    overall & 6.54 & \textbf{6.17} & 5.81 & \textbf{5.48} & 1.37 & \textbf{1.17} & 2.72 & \textbf{2.44} & 29.3 & \textbf{70.7} & 0.953 & \textbf{0.956} \\\bottomrule
    \end{tabular}
    }
    \caption{\footnotesize Quantitative results of our main experiment for evaluating the full VR telepresence system robustness. MCA outperforms CA across various metrics. Please refer to the text for details.}
    \label{tab:quantitative}
\end{table}

\newcommand{\cmark}{\ding{51}}
\newcommand{\xmark}{\ding{55}}
\begin{table}[!t]
    \centering
    \setlength{\tabcolsep}{3pt}
    {\scriptsize
    \begin{tabular}{x{14mm}|x{14mm}|x{14mm}|x{14mm}|x{14mm}|x{14mm}?{.1em}x{11mm}|x{8mm}}
    \toprule
    \textbf{blend} & \textbf{end2end} & \textbf{soft-ex.} & \textbf{dimen.} & \textbf{skip-mod.} & \textbf{tconv.} & \textbf{RMSE} & $\mathbf{\Delta}_\downarrow$\\\specialrule{.1em}{.05em}{.05em}
    - & - & - & - & - & - & 7.33 & -\\\hline
    \checkmark & - & - & - & - & - & 6.67 & 0.66\\\hline
    \checkmark & \checkmark & - & - & - & - & 6.10 & 0.57\\\hline
    \checkmark & \checkmark & \checkmark & - & - & - & 5.71 & 0.39\\\hline
    \checkmark & \checkmark & \checkmark & \checkmark & - & - & 5.64 & 0.07\\\hline
    \checkmark & \checkmark & \checkmark & \checkmark & \checkmark & - & 5.52 & 0.08\\\hline
    \checkmark & \checkmark & \checkmark & \checkmark & \checkmark & \checkmark & 5.48 & 0.04\\
    \bottomrule
    \end{tabular}
    }
    \caption{\footnotesize An ablation study of our method showing the progression and contribution to the overall performance improvement on the main RMSE metric.}
    \label{tab:ablation}
\end{table}

\begin{figure}[t!]
    \centering
    \setlength{\tabcolsep}{1pt}
    \tabulinesep=\tabcolsep
    \begin{tabu}  to \textwidth {l?{.1em}X[c,m]|X[c,m]|X[c,m]?{.1em}X[c,m]|X[c,m]|X[c,m]}
    \specialrule{.1em}{.05em}{.05em}
    \rotatebox[origin=c]{90}{{\scriptsize person1}} &
    \includegraphics[width=0.12\textwidth]{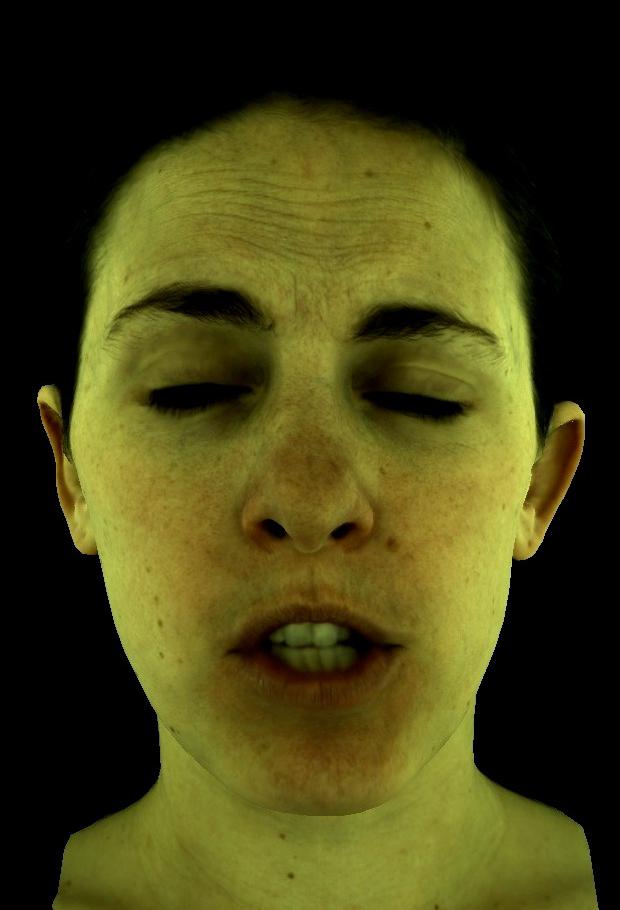} & 
    \includegraphics[width=0.12\textwidth]{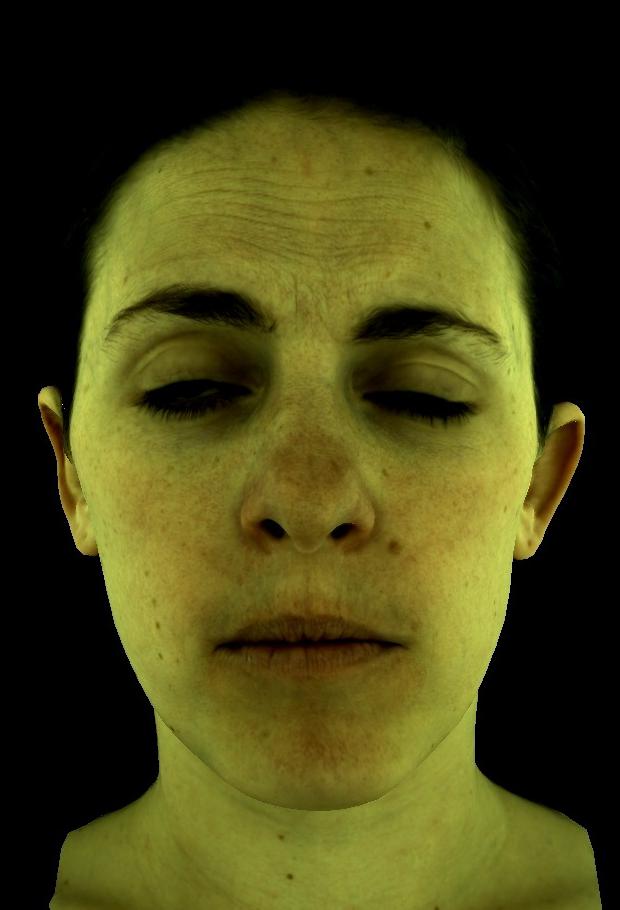} &
    \includegraphics[width=0.12\textwidth]{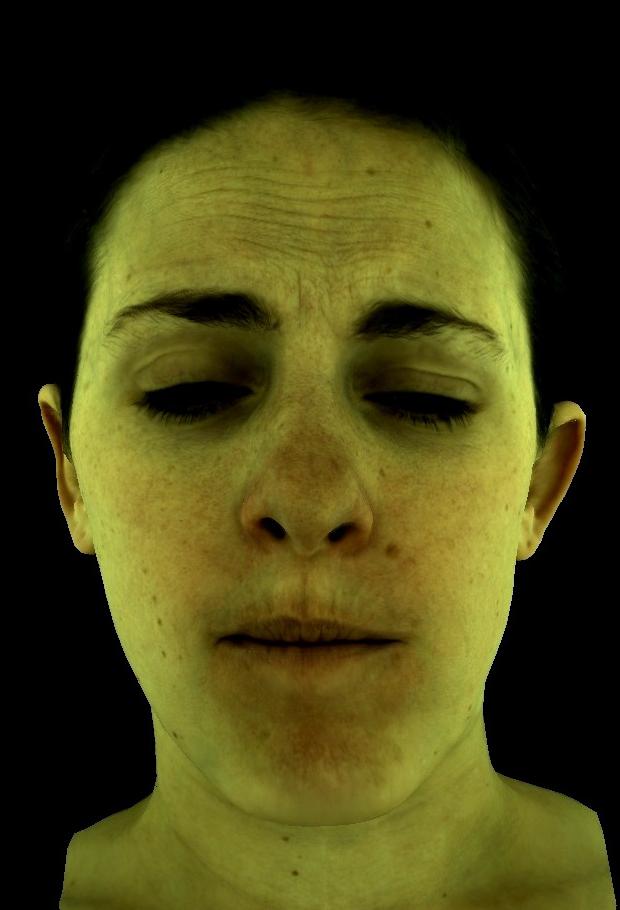} &
    \includegraphics[width=0.12\textwidth]{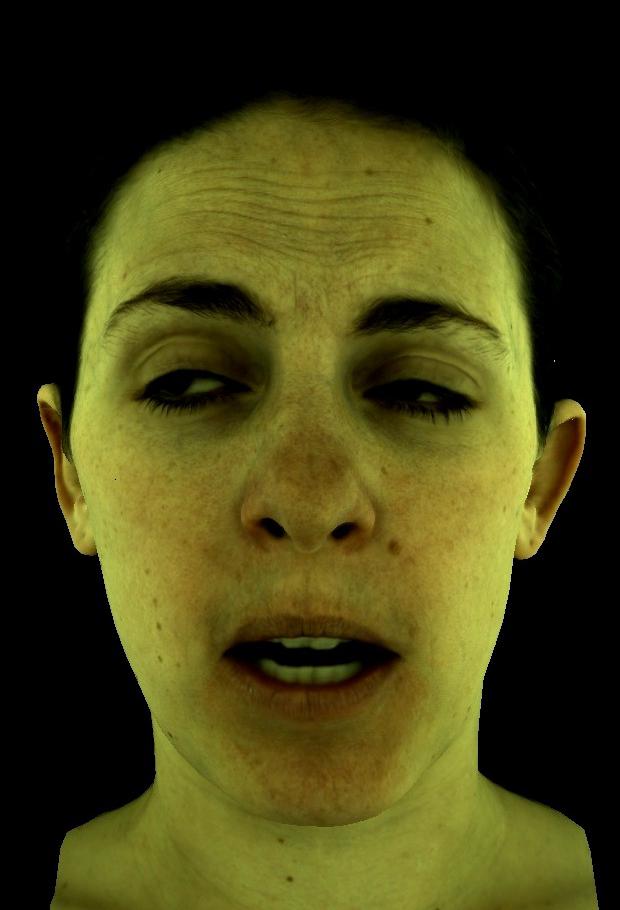} &
    \includegraphics[width=0.12\textwidth]{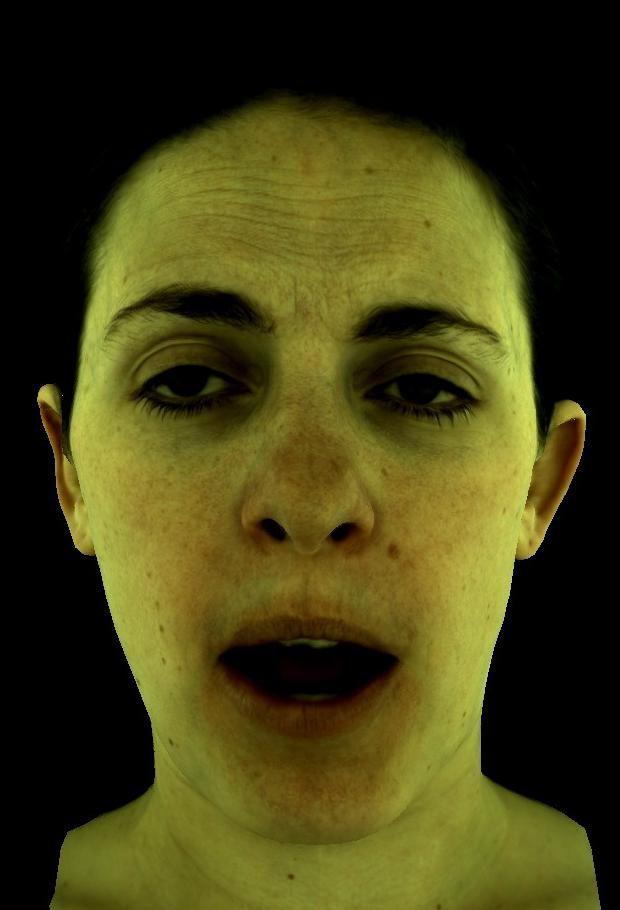} &
    \includegraphics[width=0.12\textwidth]{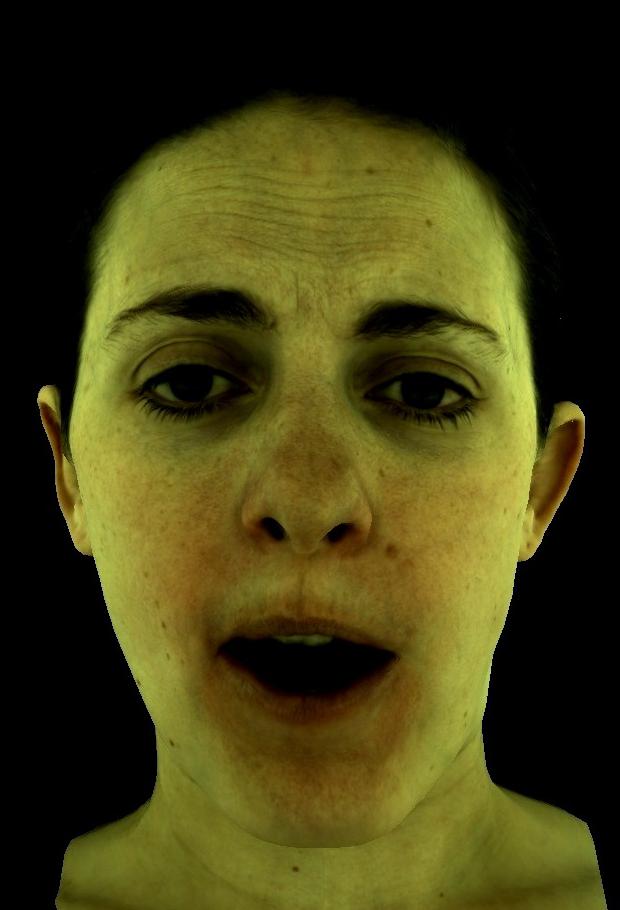}\\\hline
    \rotatebox[origin=c]{90}{{\scriptsize person2}} &
    \includegraphics[width=0.12\textwidth]{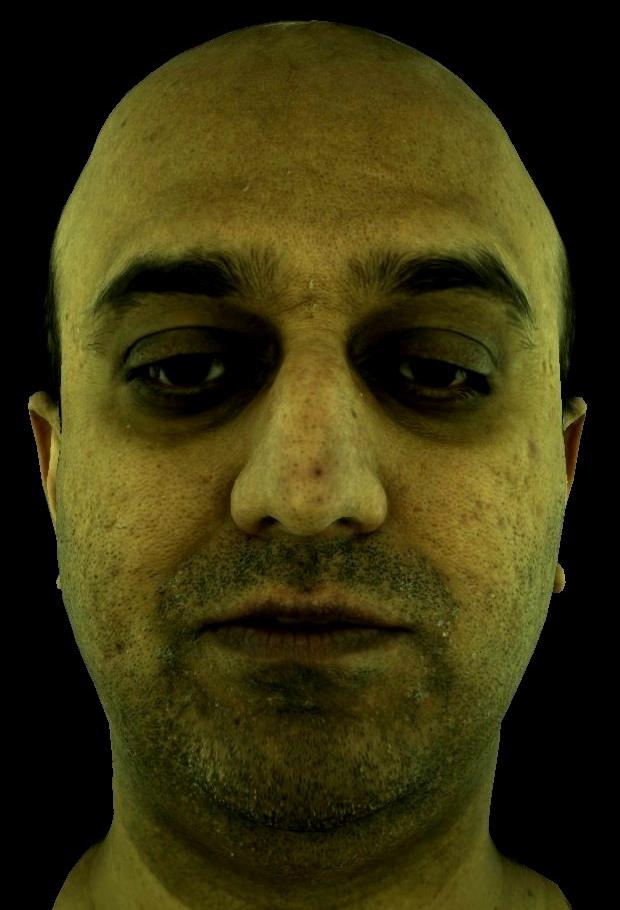} & 
    \includegraphics[width=0.12\textwidth]{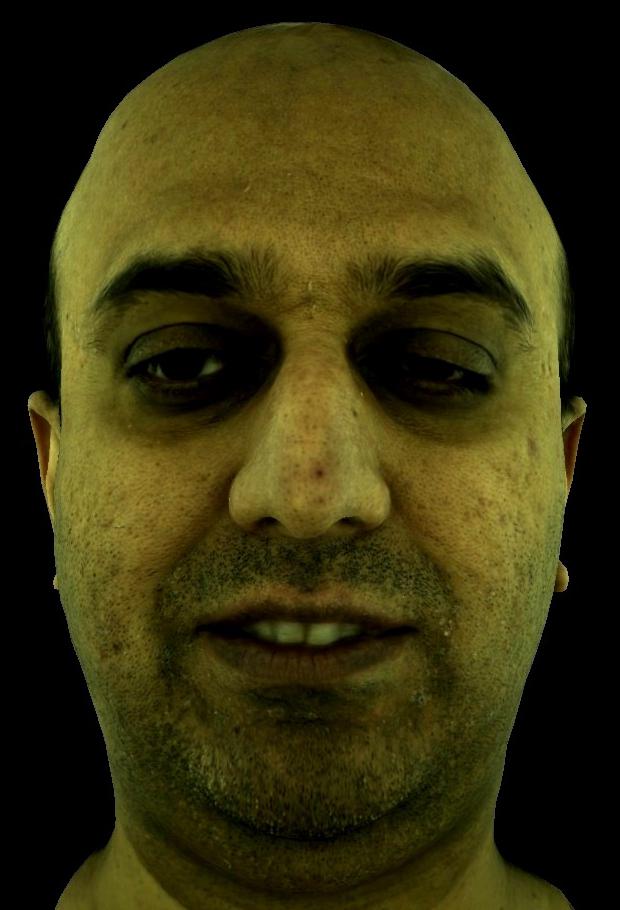} &
    \includegraphics[width=0.12\textwidth]{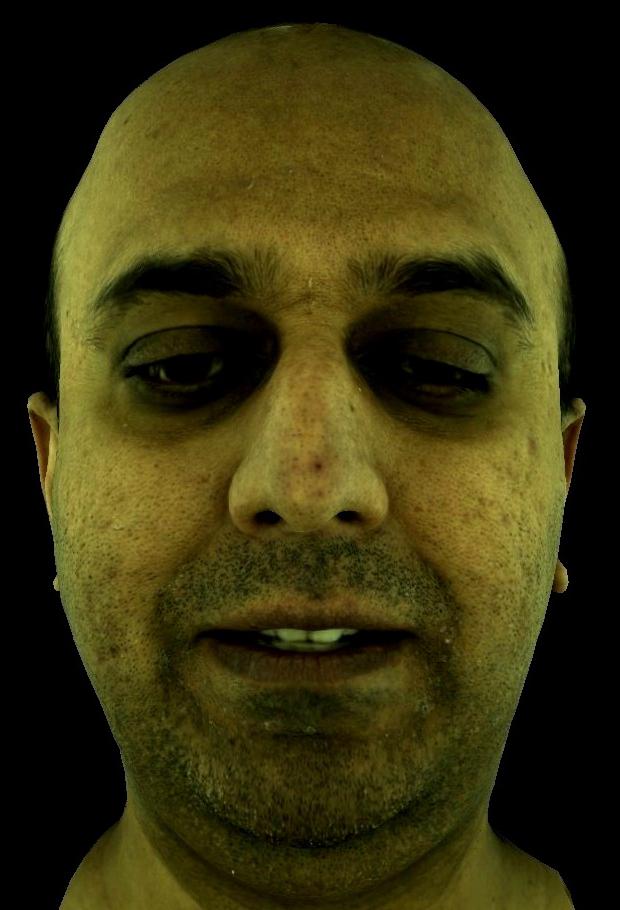} &
    \includegraphics[width=0.12\textwidth]{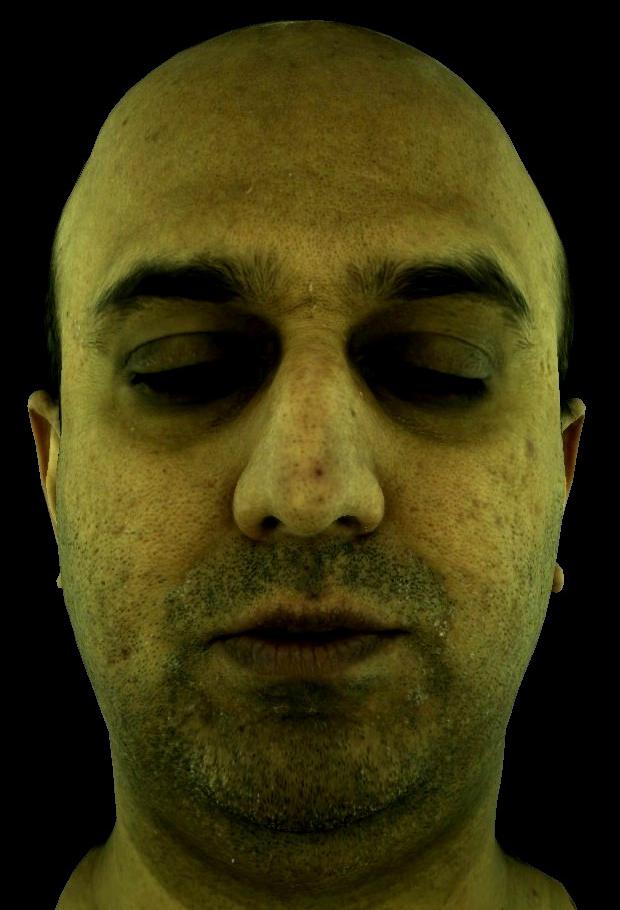} &
    \includegraphics[width=0.12\textwidth]{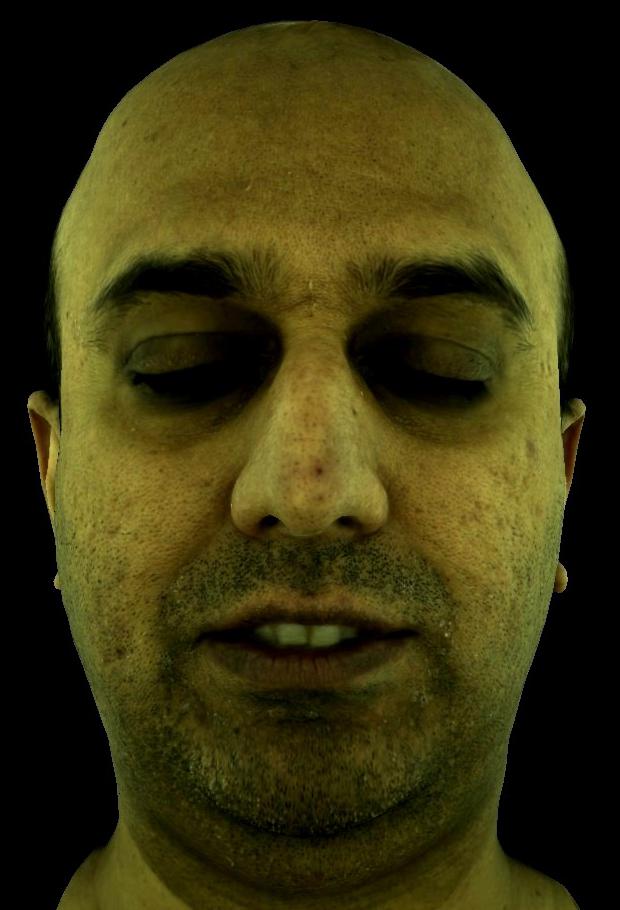} &
    \includegraphics[width=0.12\textwidth]{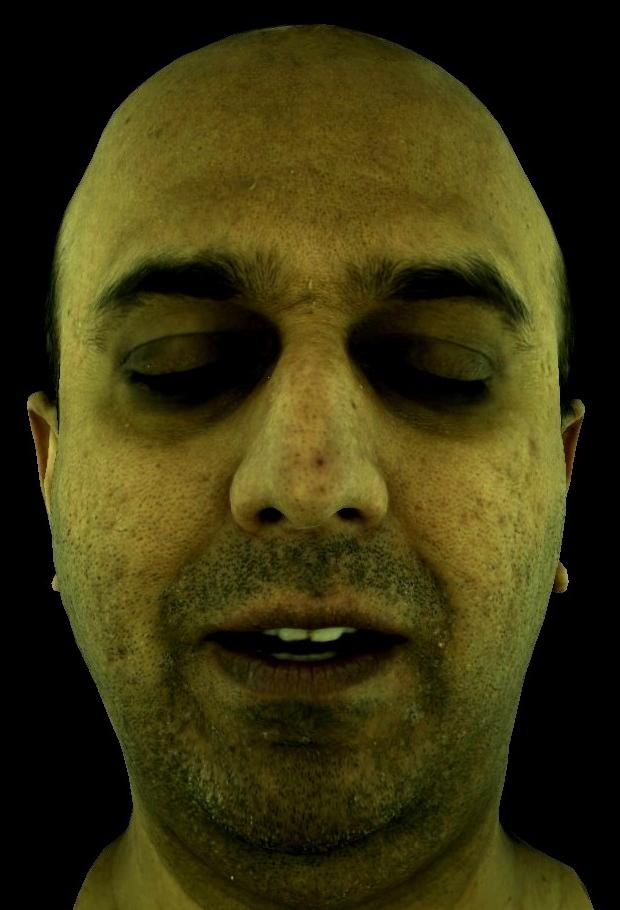}\\\hline
    \rotatebox[origin=c]{90}{{\scriptsize person3}} &
    \includegraphics[width=0.12\textwidth]{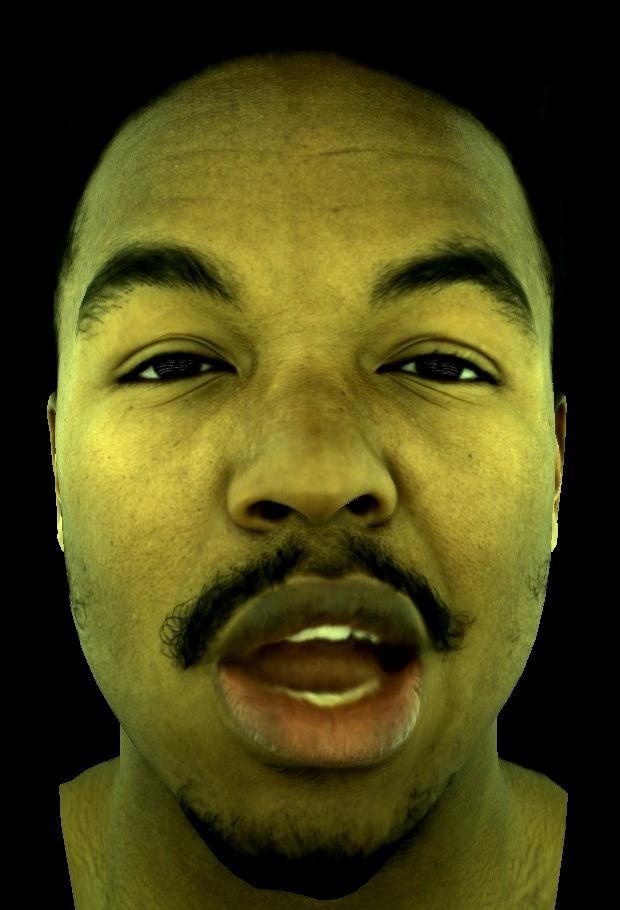} & 
    \includegraphics[width=0.12\textwidth]{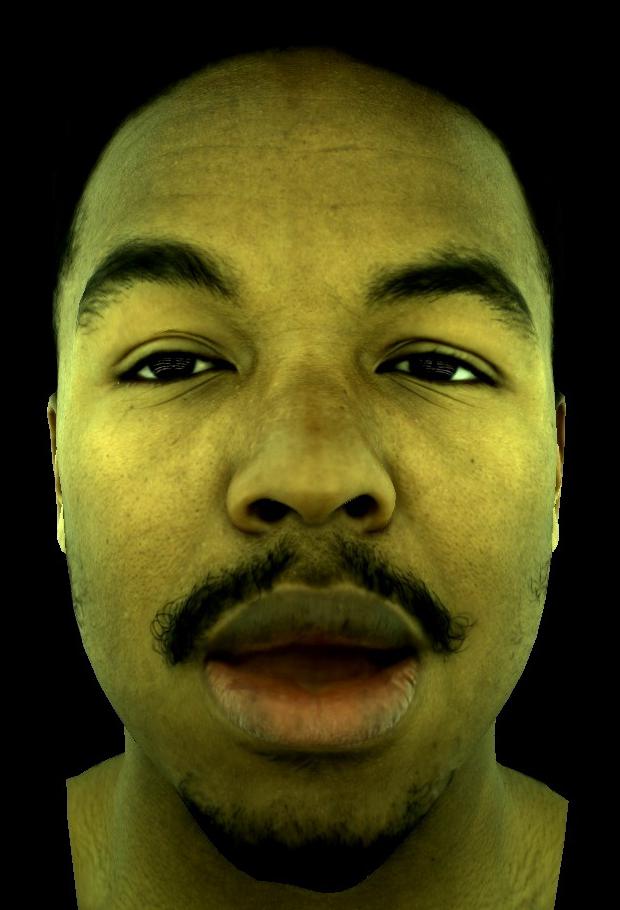} &
    \includegraphics[width=0.12\textwidth]{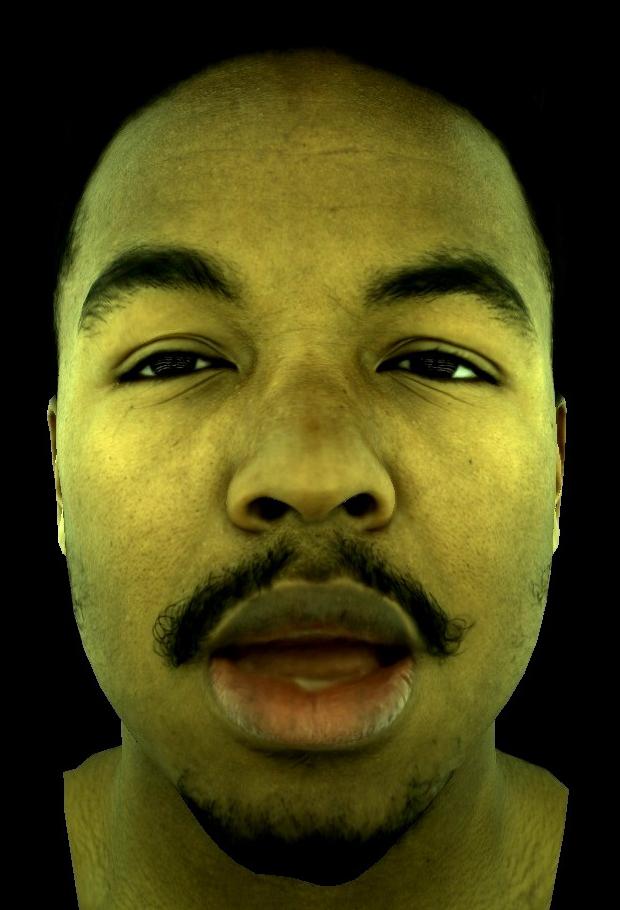} &
    \includegraphics[width=0.12\textwidth]{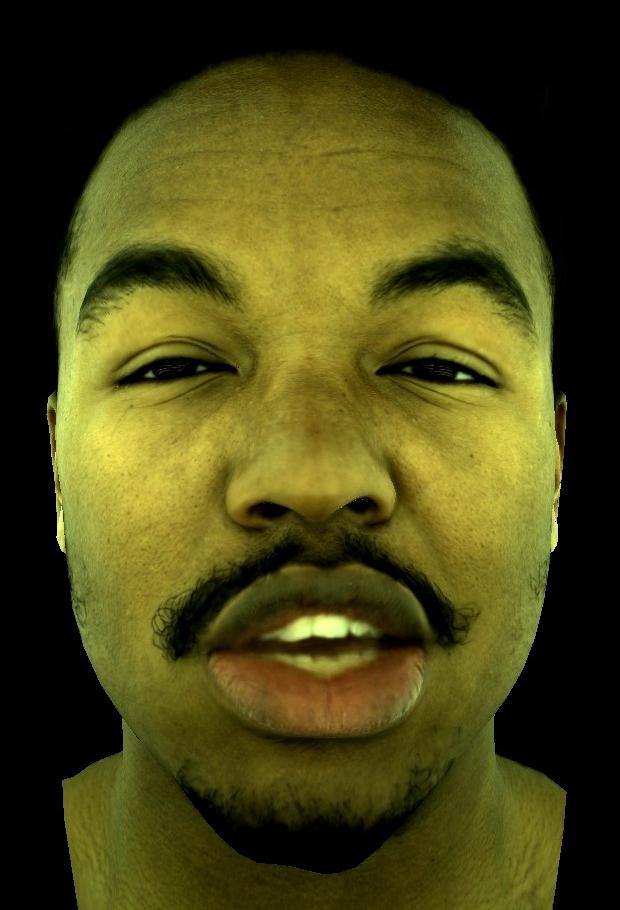} &
    \includegraphics[width=0.12\textwidth]{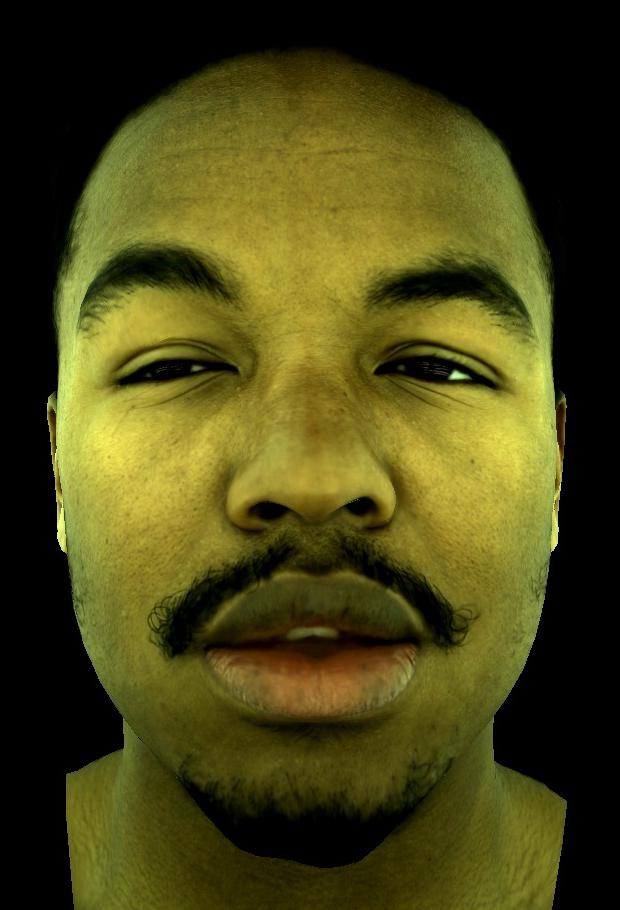} &
    \includegraphics[width=0.12\textwidth]{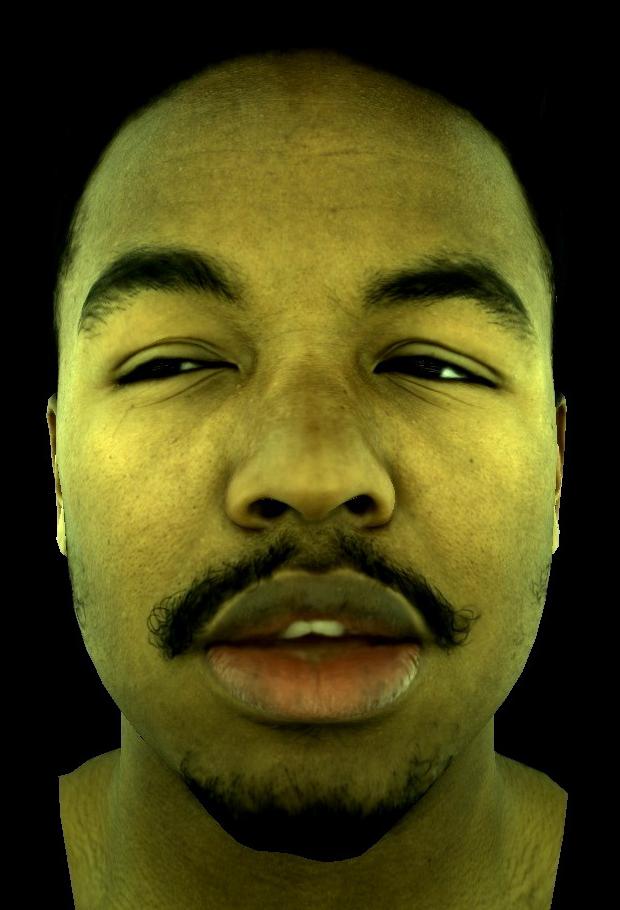}\\\hline
    \rotatebox[origin=c]{90}{{\scriptsize person4}} &
    \includegraphics[width=0.12\textwidth]{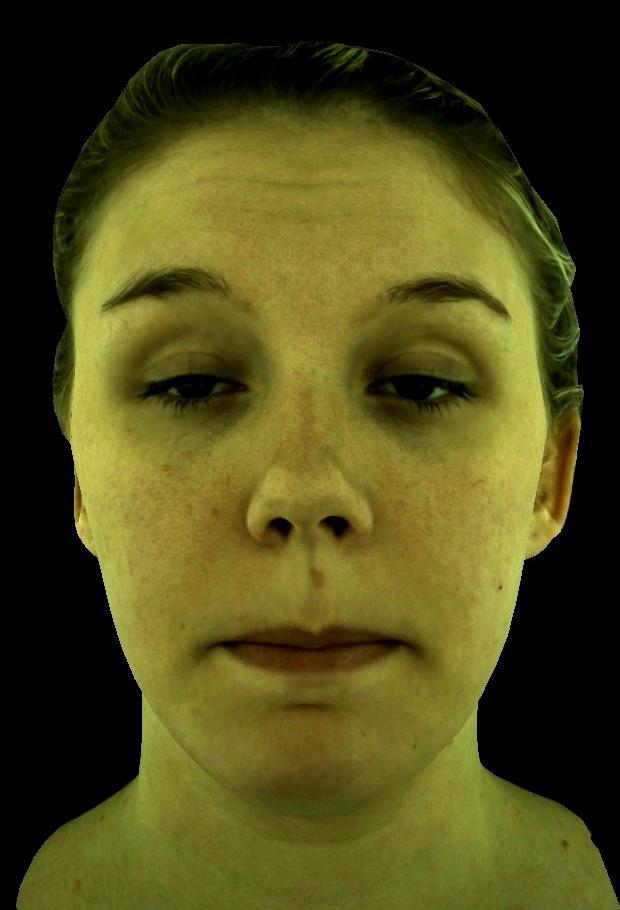} & 
    \includegraphics[width=0.12\textwidth]{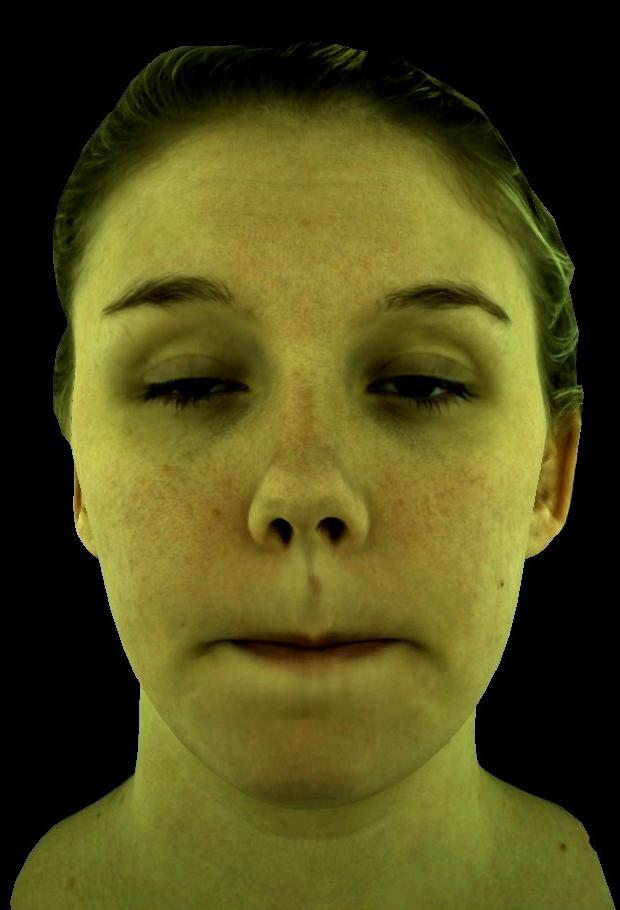} &
    \includegraphics[width=0.12\textwidth]{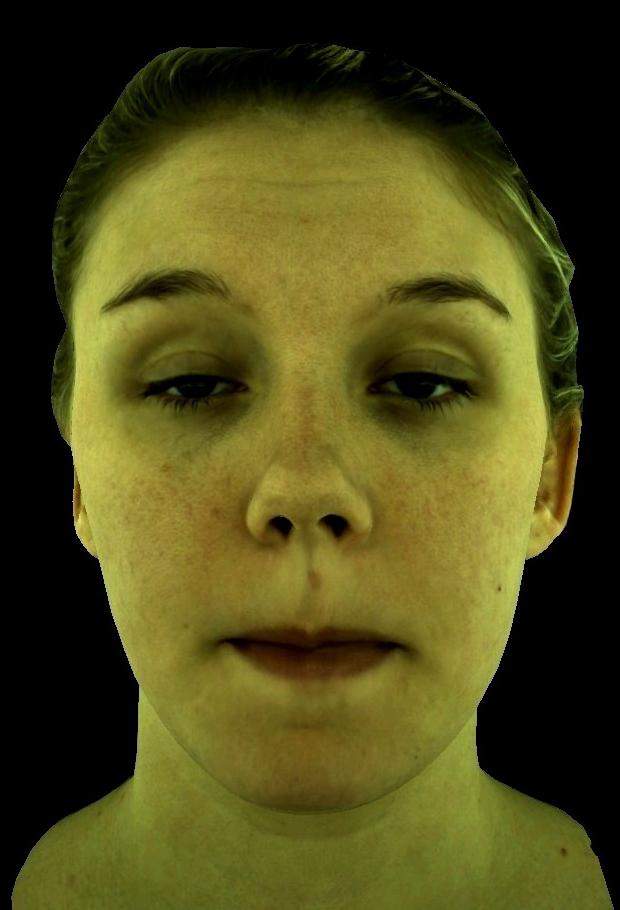} &
    \includegraphics[width=0.12\textwidth]{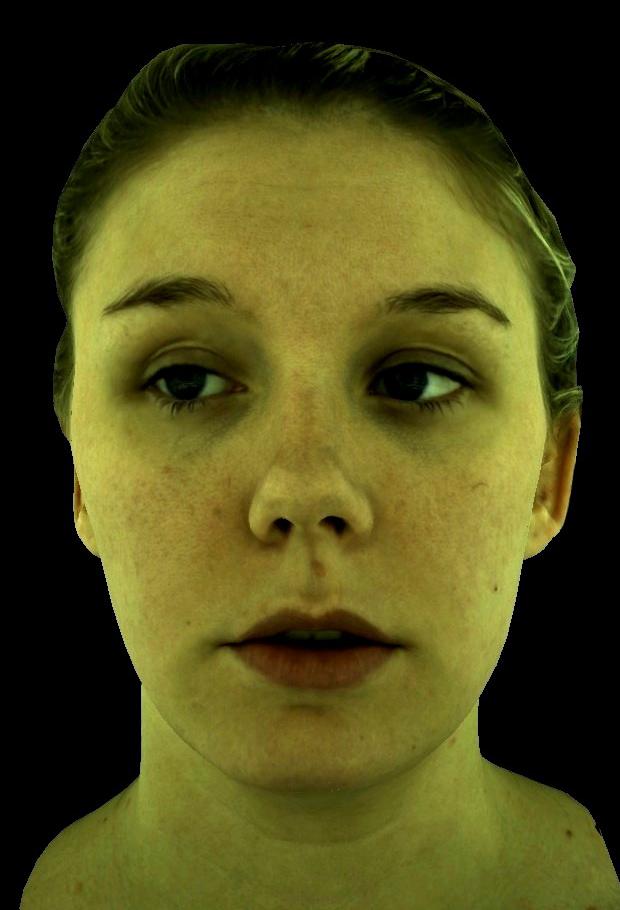} &
    \includegraphics[width=0.12\textwidth]{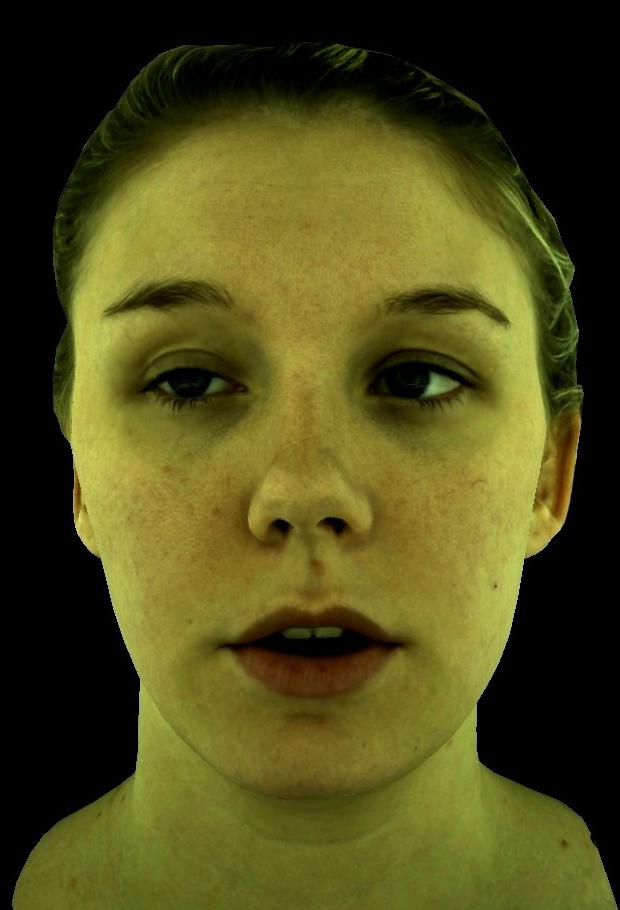} &
    \includegraphics[width=0.12\textwidth]{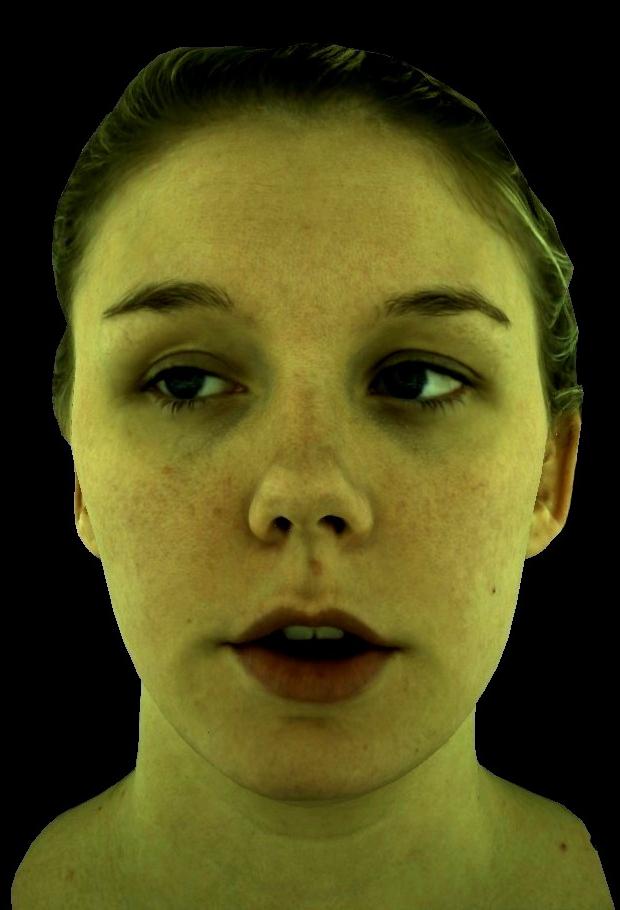}\\\specialrule{.1em}{.05em}{.05em} 
    ~ & \scriptsize{CA~\cite{wei2019vr,lombardi2018deep}} & \scriptsize{MCA(ours)} & \scriptsize{GT} & \scriptsize{CA~\cite{wei2019vr,lombardi2018deep}} & \scriptsize{MCA(ours)} & \scriptsize{GT}\\
    \specialrule{.1em}{.05em}{.05em} 
    \end{tabu}
    \caption{\footnotesize Qualitative VR telepresence results. Compared to the holistic approach (CA), MCA handles untrained subtle expressions better.}
    \label{fig:qualitative}
\end{figure}

\begin{figure}[t!]
    \centering
    \setlength{\tabcolsep}{1pt}
    \tabulinesep=\tabcolsep
    \begin{tabu} to \textwidth {X[c,m]|X[c,m]|X[c,m]?{.1em}X[c,m]|X[c,m]|X[c,m]}
    \specialrule{.1em}{.05em}{.05em}
    \includegraphics[width=0.12\textwidth]{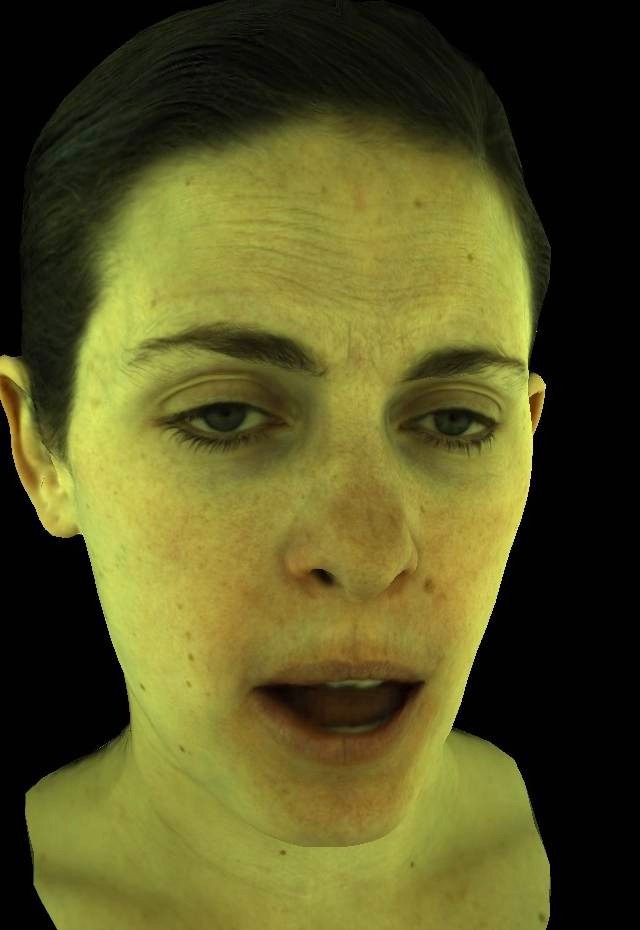} & 
    \includegraphics[width=0.12\textwidth]{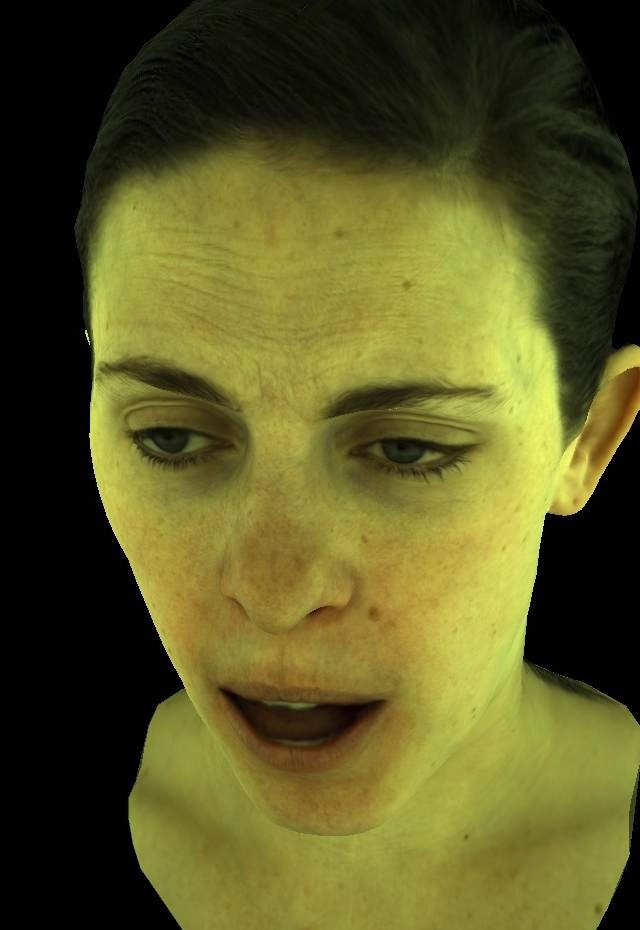} &
    \includegraphics[width=0.12\textwidth]{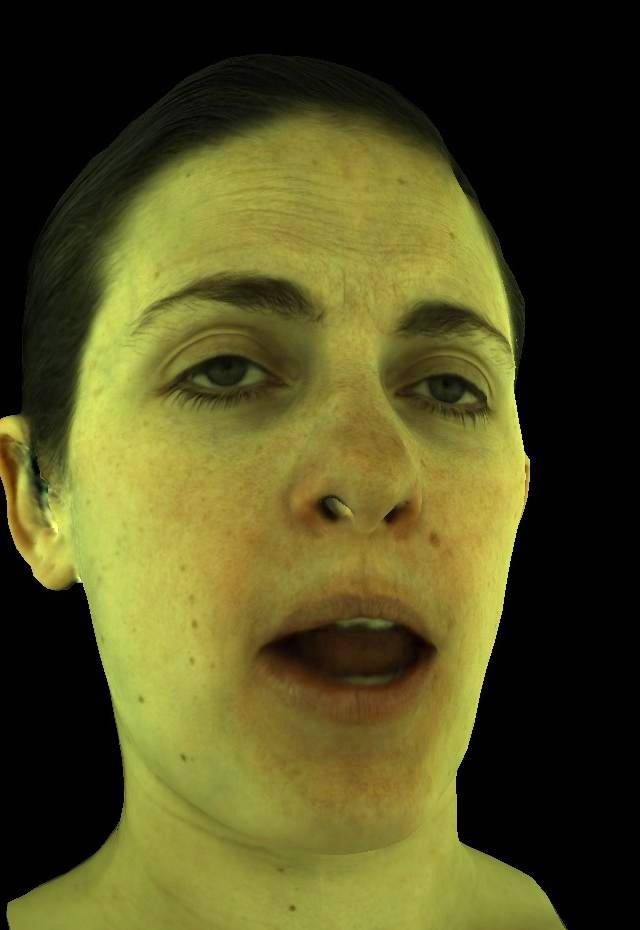} &
    \includegraphics[width=0.12\textwidth]{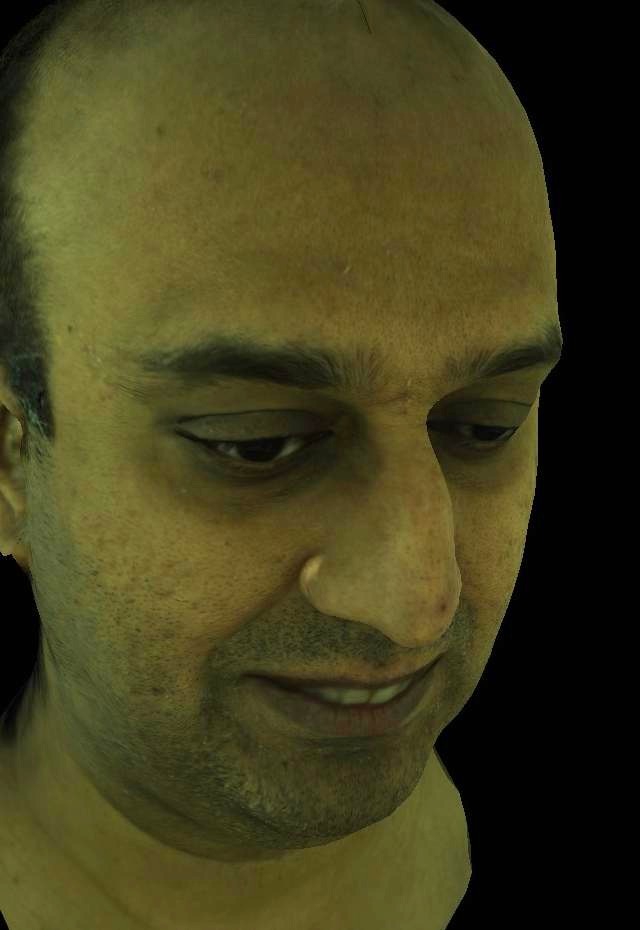} &
    \includegraphics[width=0.12\textwidth]{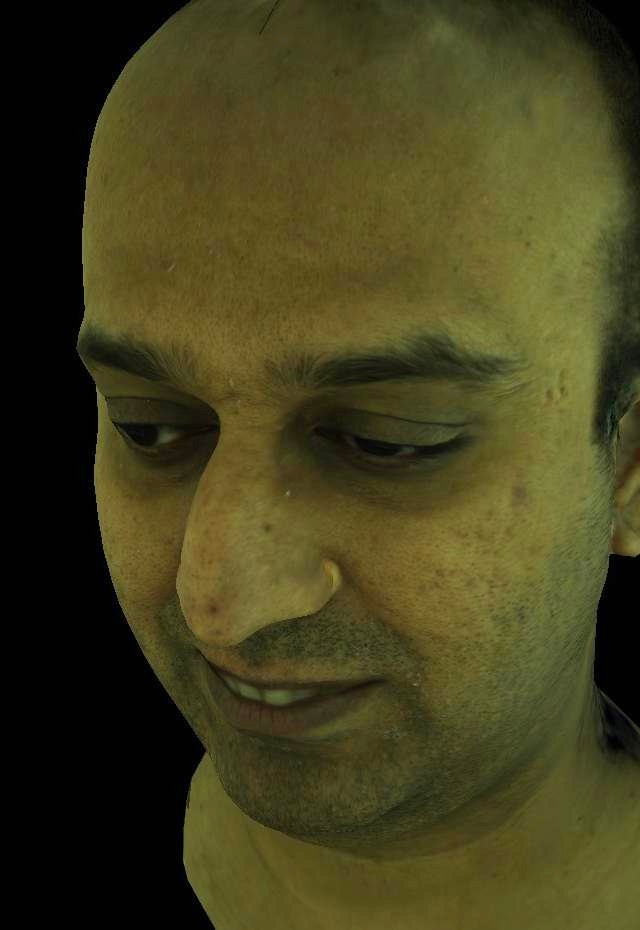} &
    \includegraphics[width=0.12\textwidth]{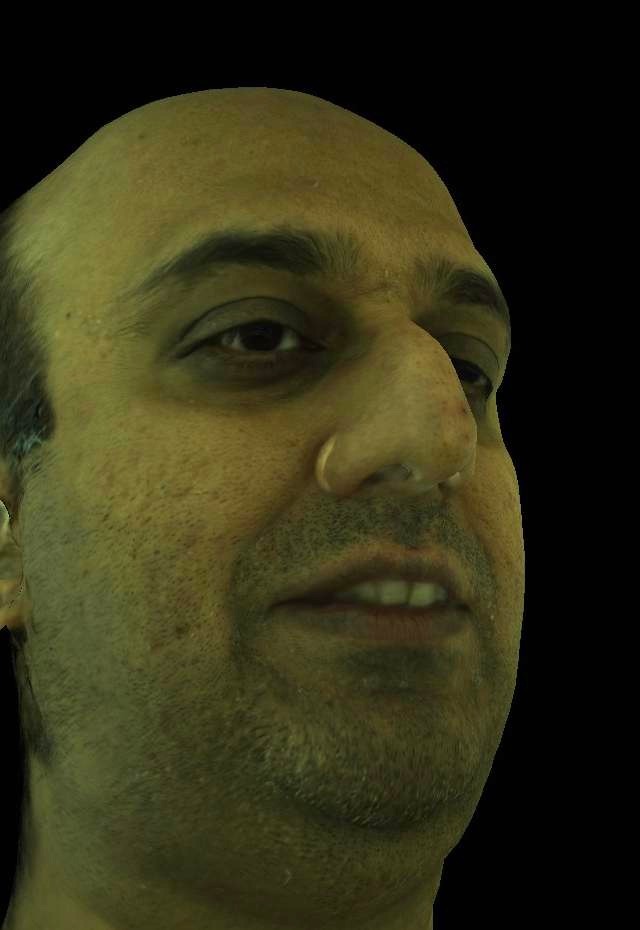}\\\hline
    \includegraphics[width=0.12\textwidth]{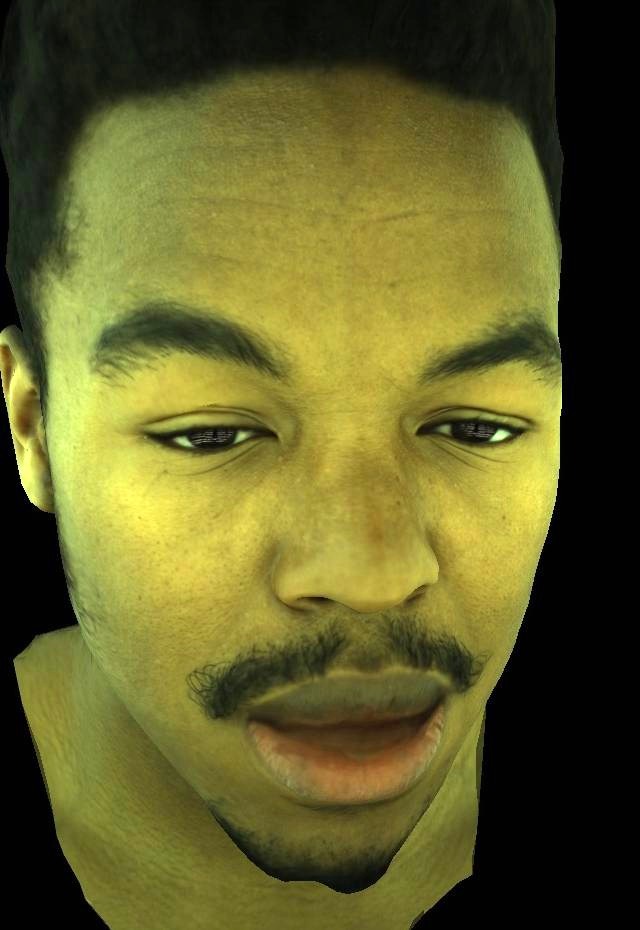} & 
    \includegraphics[width=0.12\textwidth]{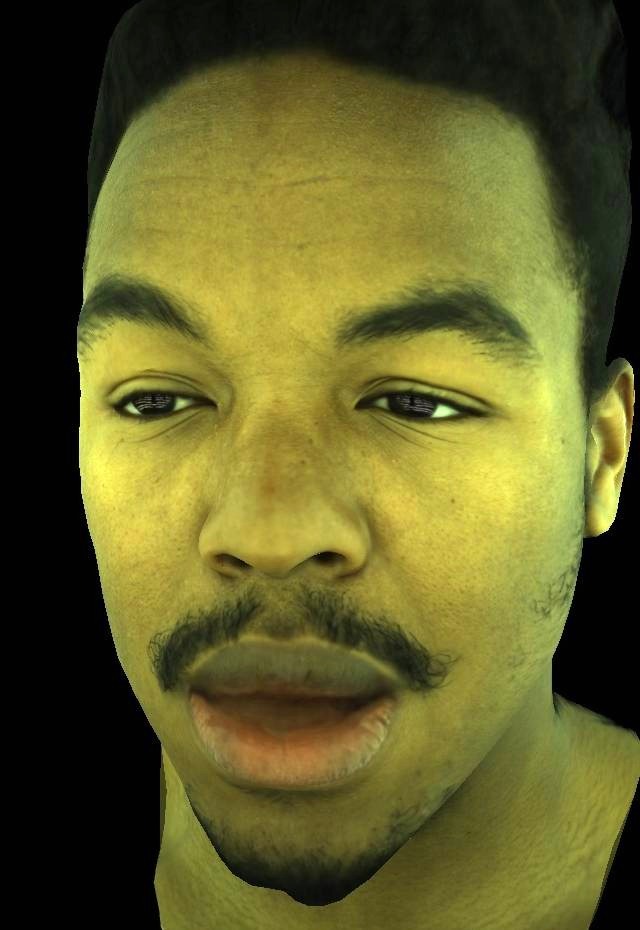} &
    \includegraphics[width=0.12\textwidth]{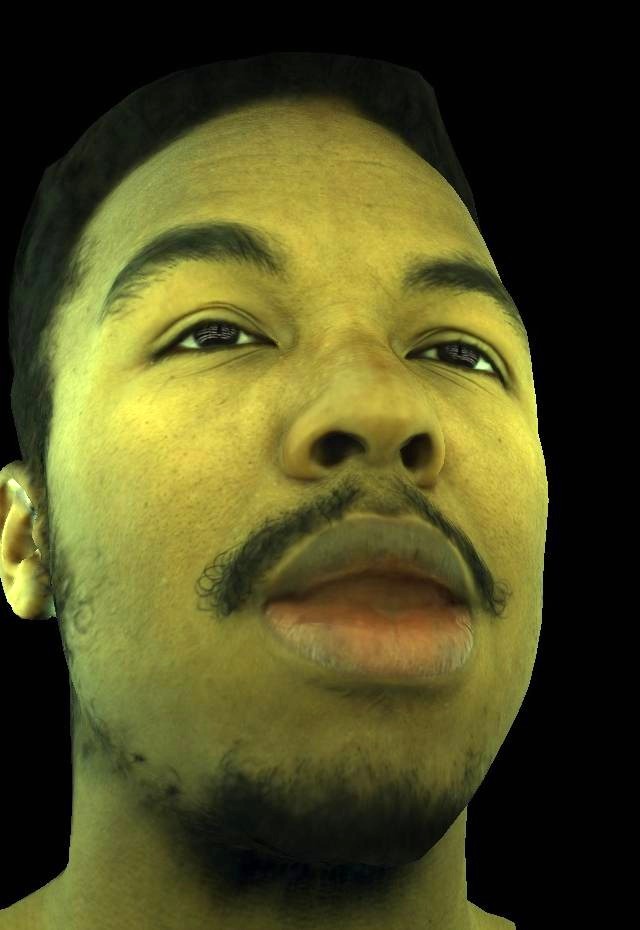} &
    \includegraphics[width=0.12\textwidth]{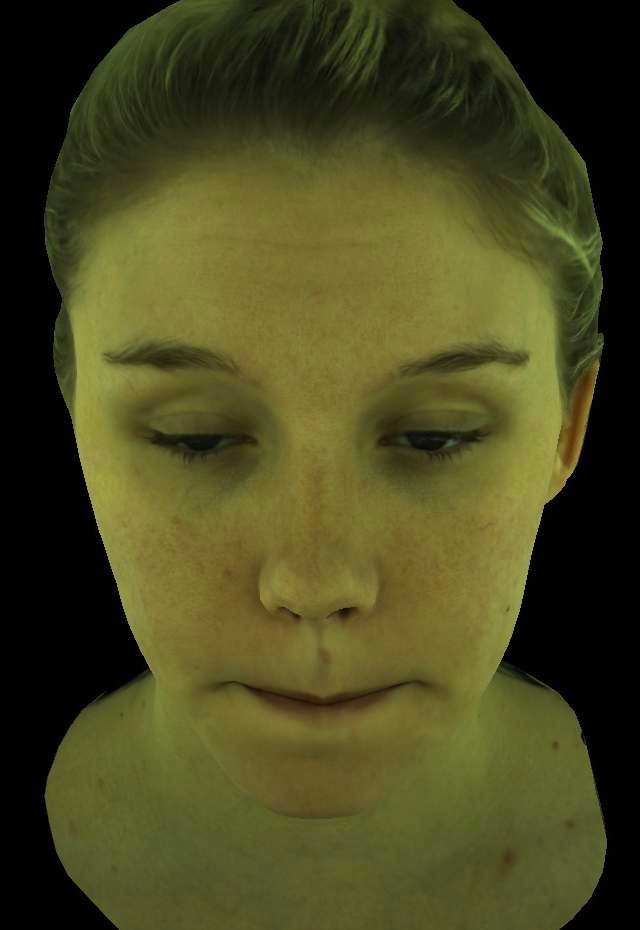} &
    \includegraphics[width=0.12\textwidth]{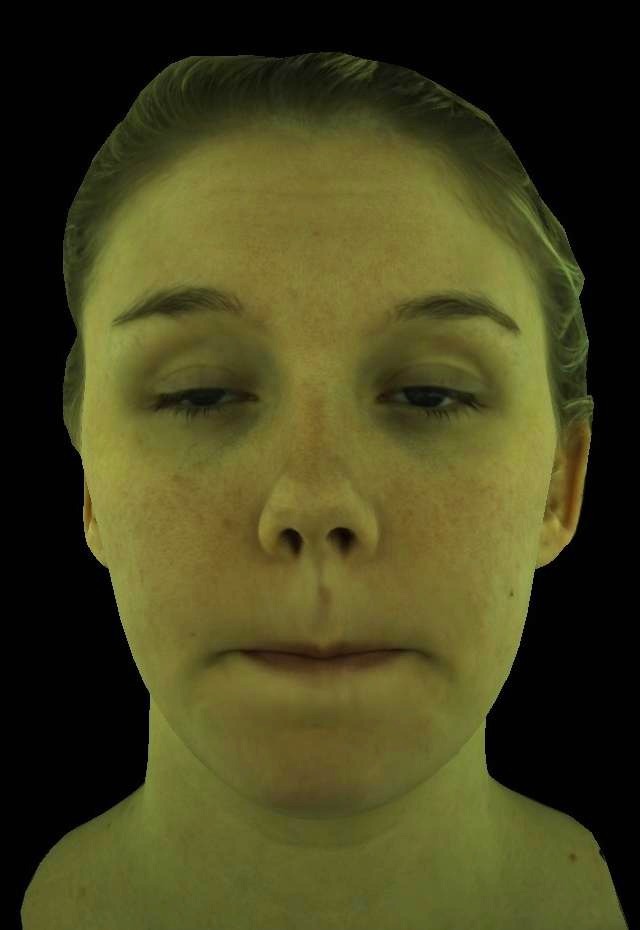} &
    \includegraphics[width=0.12\textwidth]{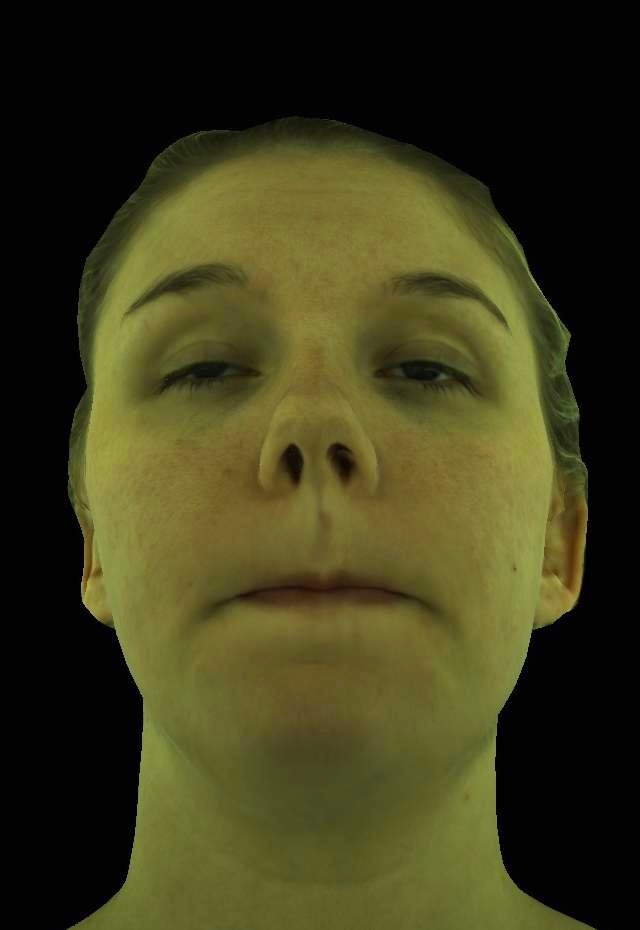}\\
    \specialrule{.1em}{.05em}{.05em}
    \end{tabu}
    \caption{\footnotesize Qualitative results of MCA from different viewing directions by varying $\mathbf{v}$. MCA produces natural expressions that are consistent across viewpoints.}
    \label{fig:viewpoint}
\end{figure}

\begin{figure}[t!]
    \centering
    \setlength{\tabcolsep}{1pt}
    \tabulinesep=\tabcolsep
    \begin{tabu} to \textwidth {X[0.5,c,m] X[0.8,c,m] X[0.8,c,m] | X[0.5,c,m] X[0.8,c,m] X[0.8,c,m]}
    \specialrule{.1em}{.05em}{.05em}
    \includegraphics[height=0.2\textwidth]{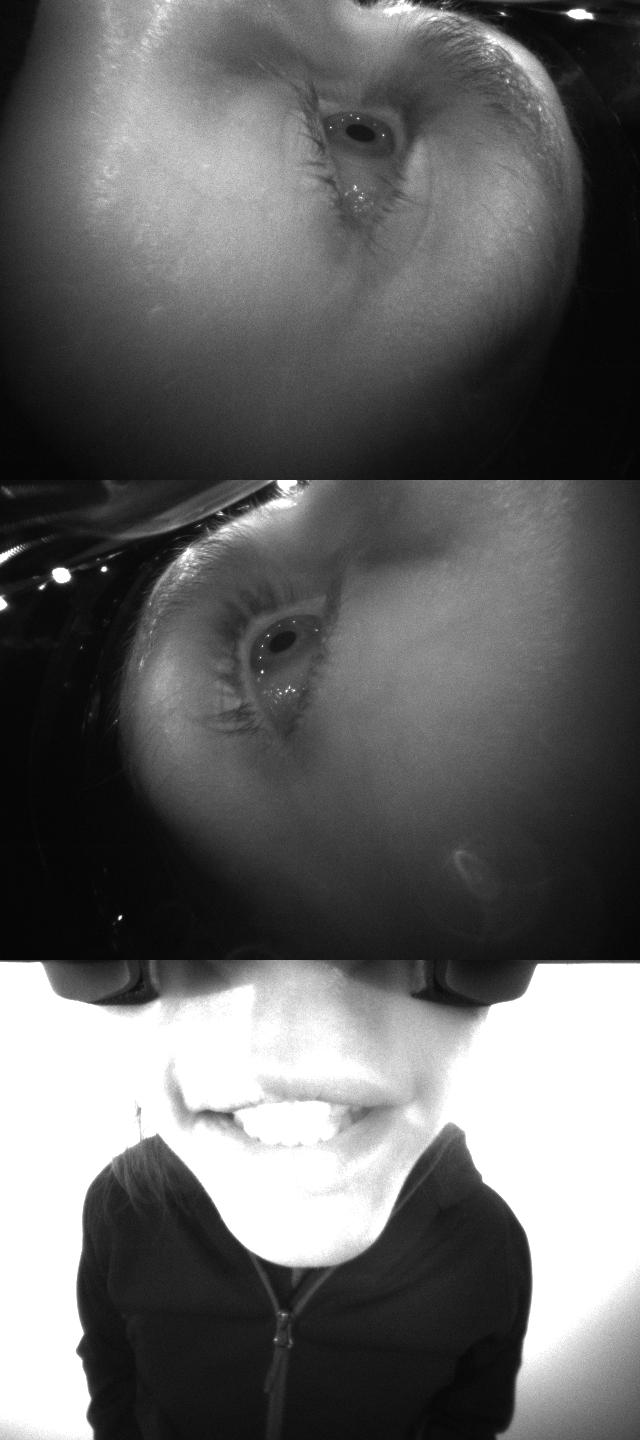} & 
    \includegraphics[height=0.2\textwidth]{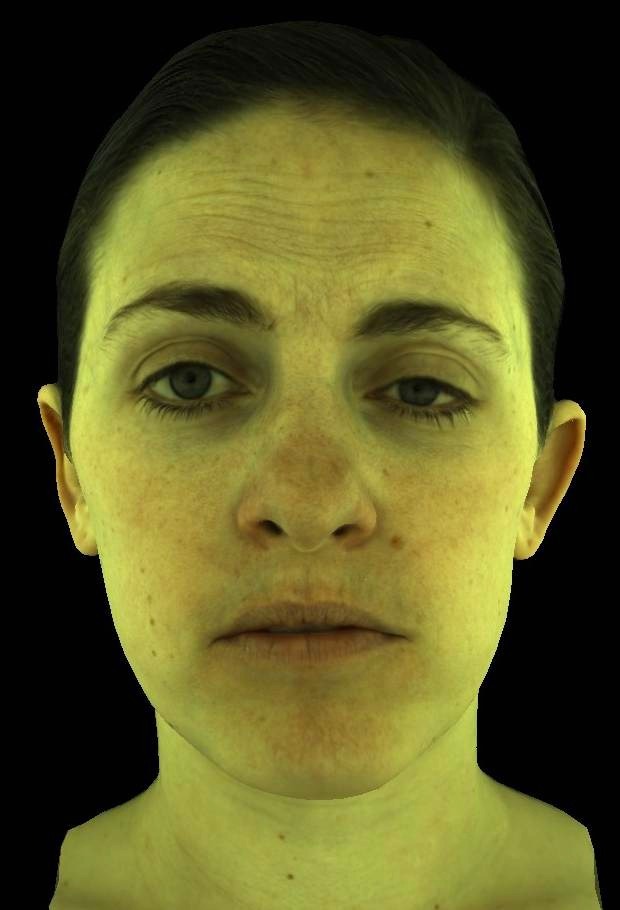} &
    \includegraphics[height=0.2\textwidth]{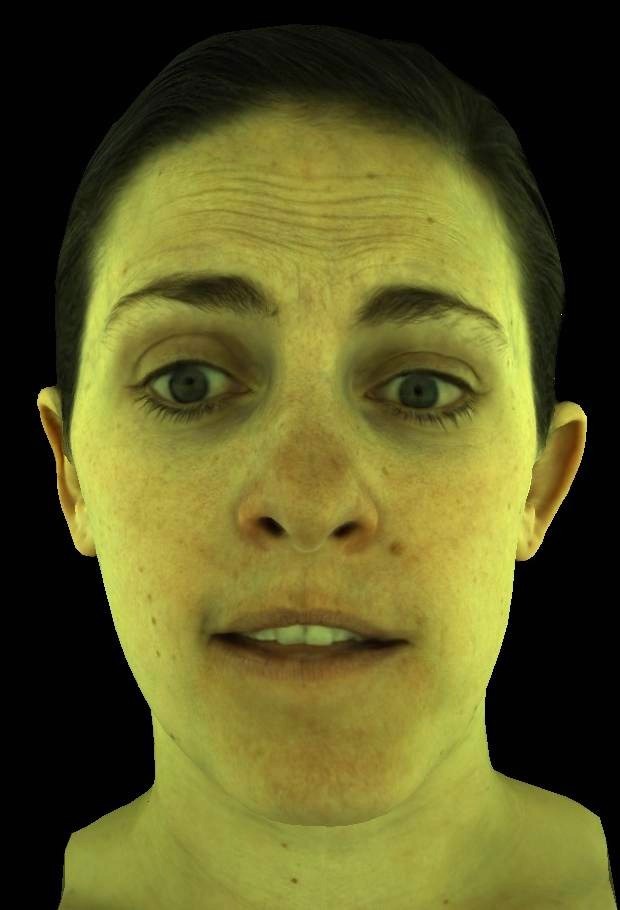} &
    \includegraphics[height=0.2\textwidth]{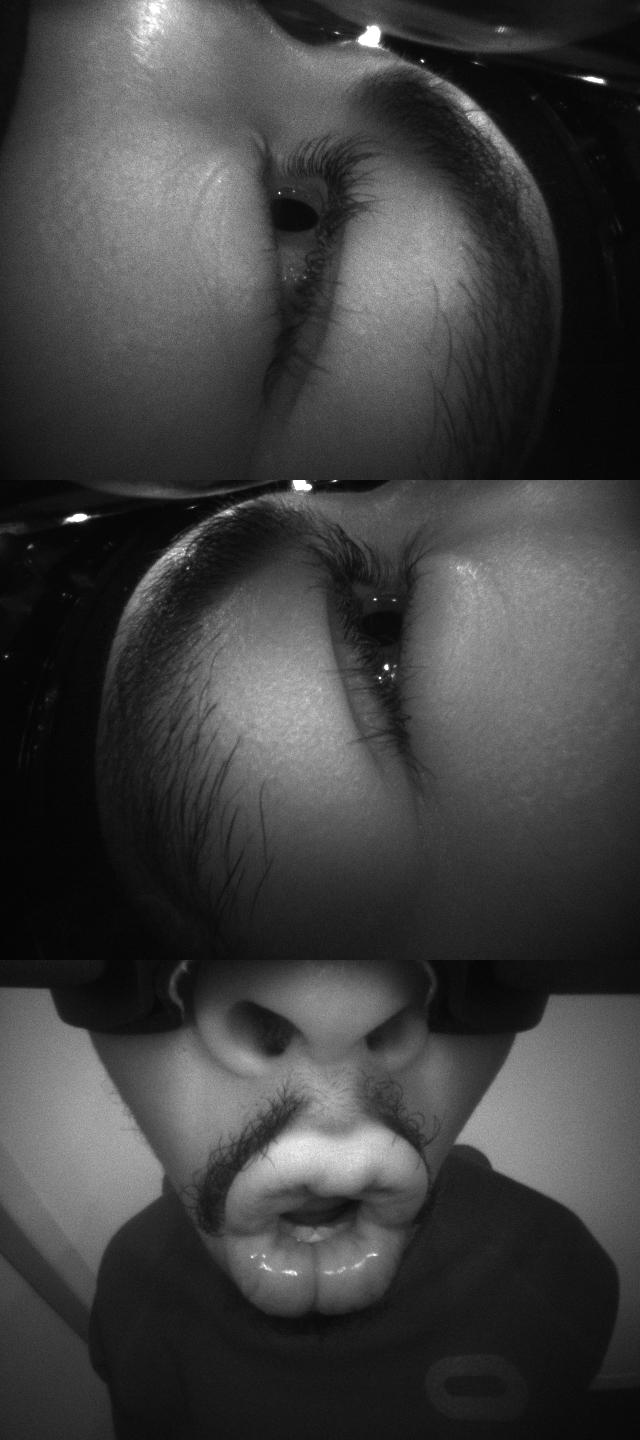} &
    \includegraphics[height=0.2\textwidth]{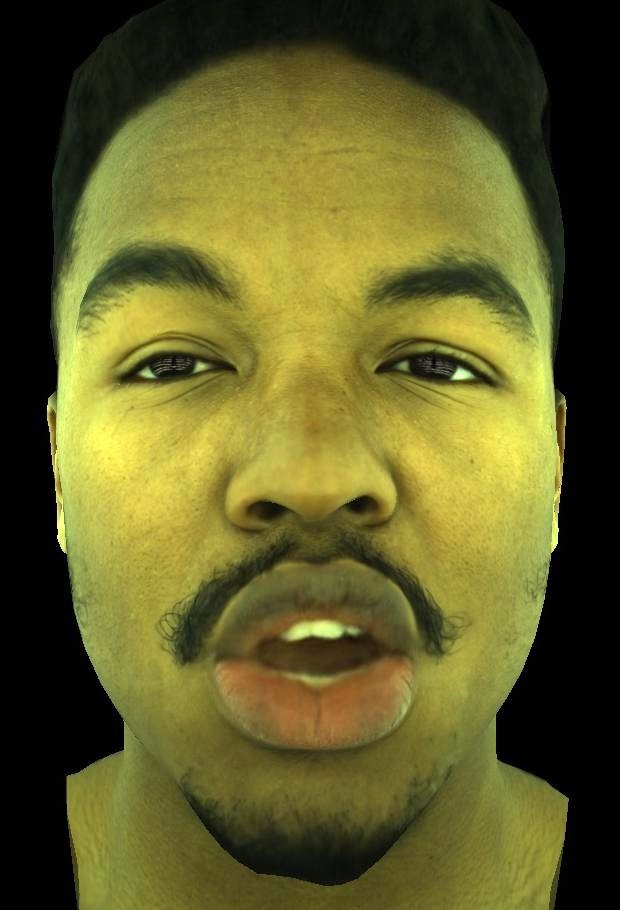} &
    \includegraphics[height=0.2\textwidth]{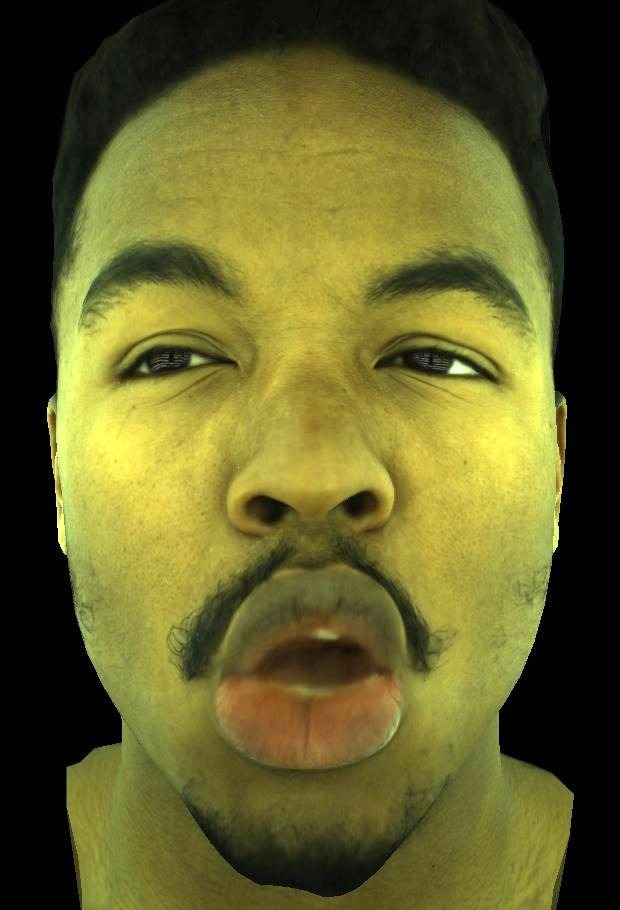}\\\hline
    \scriptsize{image} & \scriptsize{MCA} & \scriptsize{GT} & \scriptsize{image} & \scriptsize{MCA} & \scriptsize{GT}\\
    \specialrule{.1em}{.05em}{.05em}
    \end{tabu}
    \caption{\footnotesize Failure cases of MCA. Typical failure cases include interference from strong background flash, extreme asymmetry between the eyes, and weakened motion.}
    \label{fig:failure}
\end{figure}

\subsection{Extensive Applications}
\begin{figure}[t!]
    \centering
    \setlength{\tabcolsep}{1pt}
    \tabulinesep=\tabcolsep
    \begin{tabu} to \textwidth {X[0.5135,c,m] X[1.0,c,m] | X[0.5135,c,m] X[1.0,c,m] | X[1.0,c,m] X[1.0,c,m] | X[1.0,c,m] X[1.0,c,m]}
    \specialrule{.1em}{.05em}{.05em}
    \includegraphics[height=0.21\textwidth]{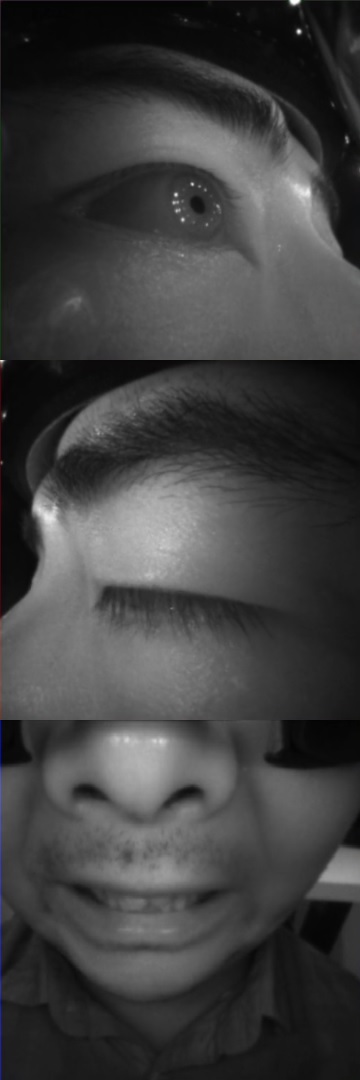} & 
    \includegraphics[height=0.21\textwidth]{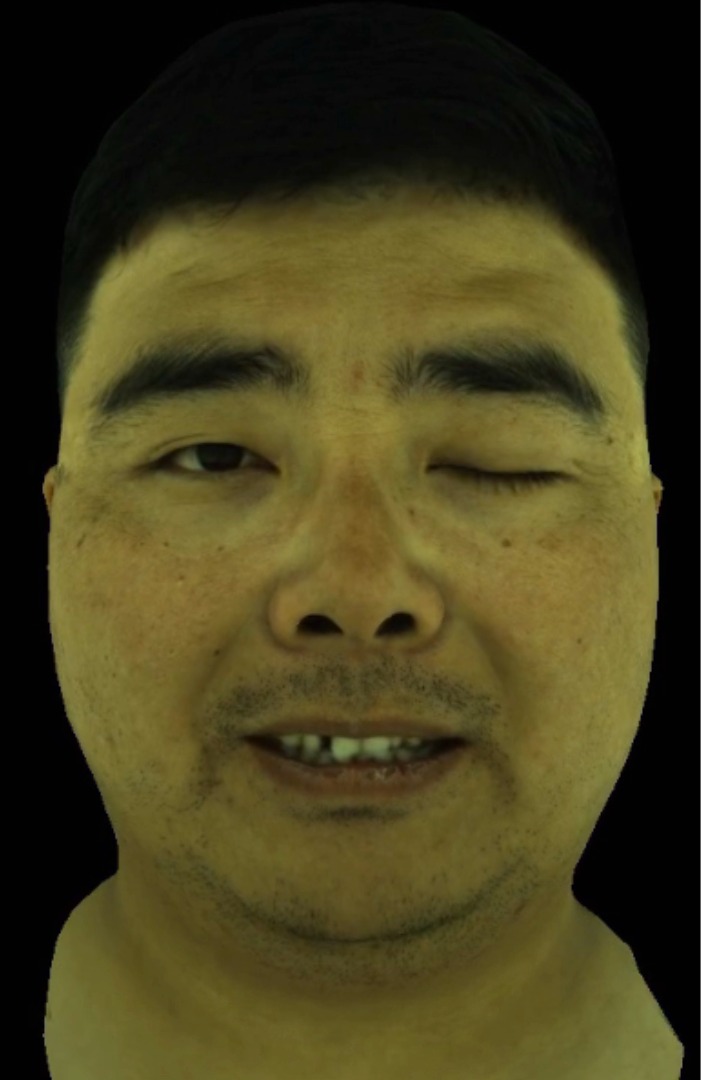} &
    \includegraphics[height=0.21\textwidth]{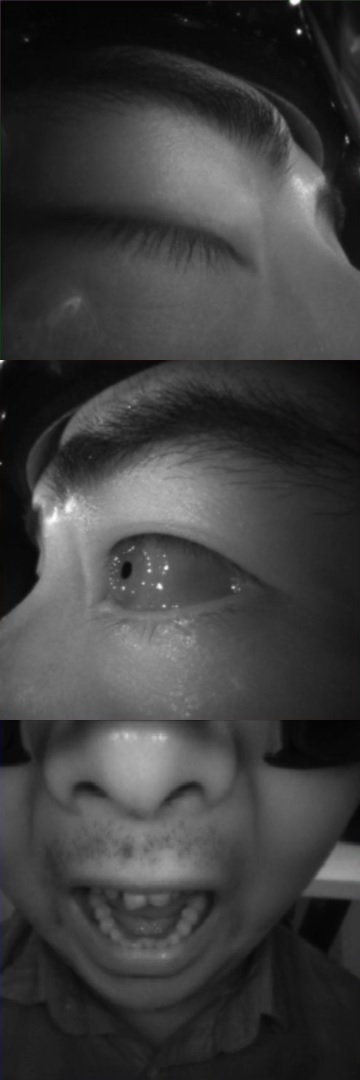} &
    \includegraphics[height=0.21\textwidth]{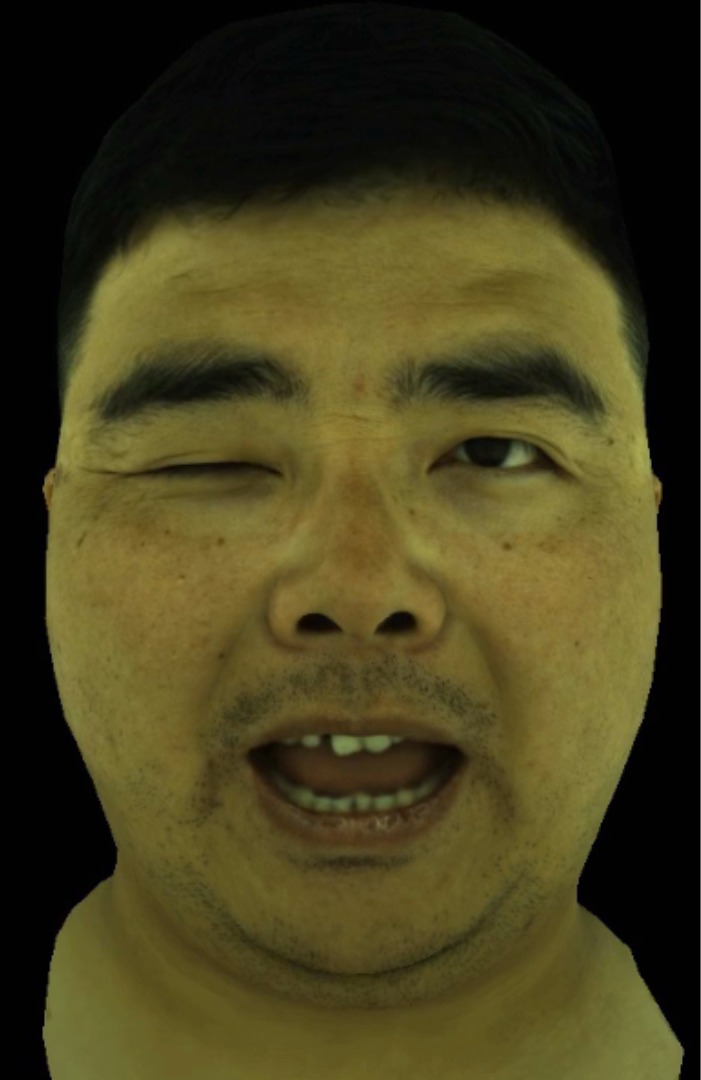} &
    \includegraphics[height=0.21\textwidth]{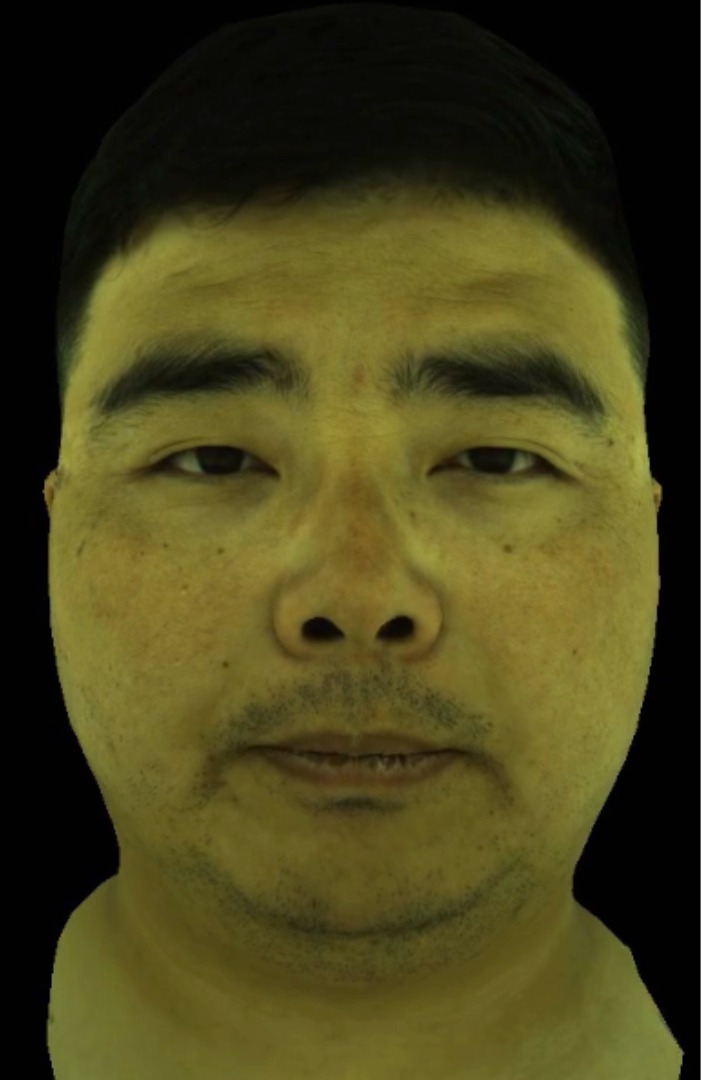} & 
    \includegraphics[height=0.21\textwidth]{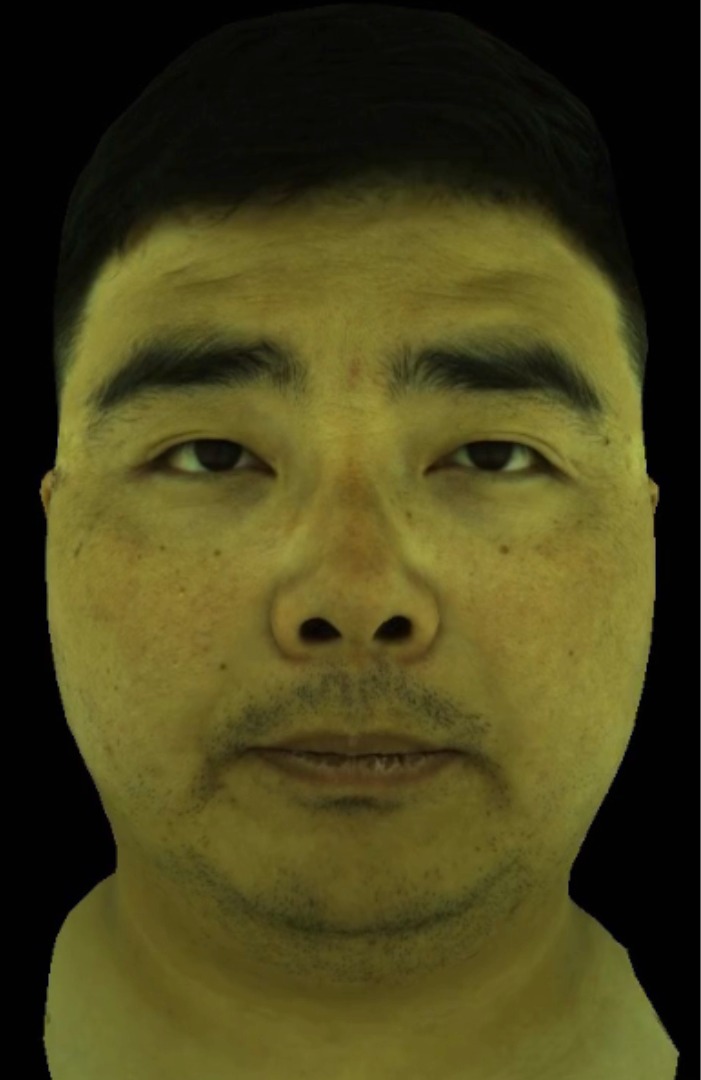} &
    \includegraphics[height=0.21\textwidth]{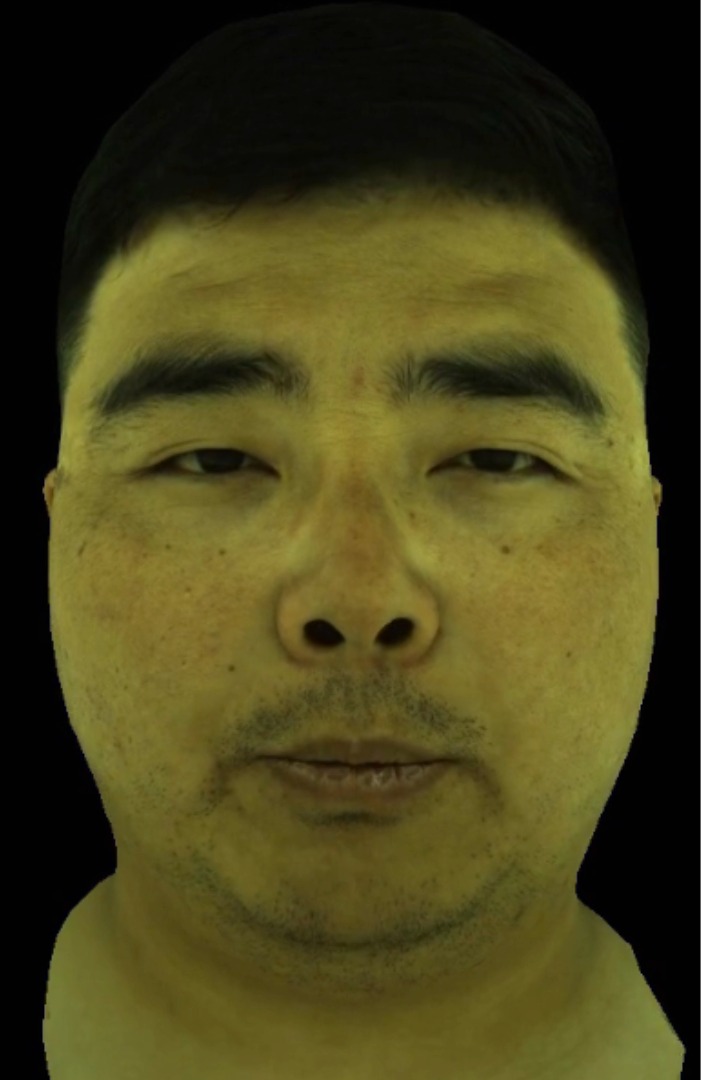} &
    \includegraphics[height=0.21\textwidth]{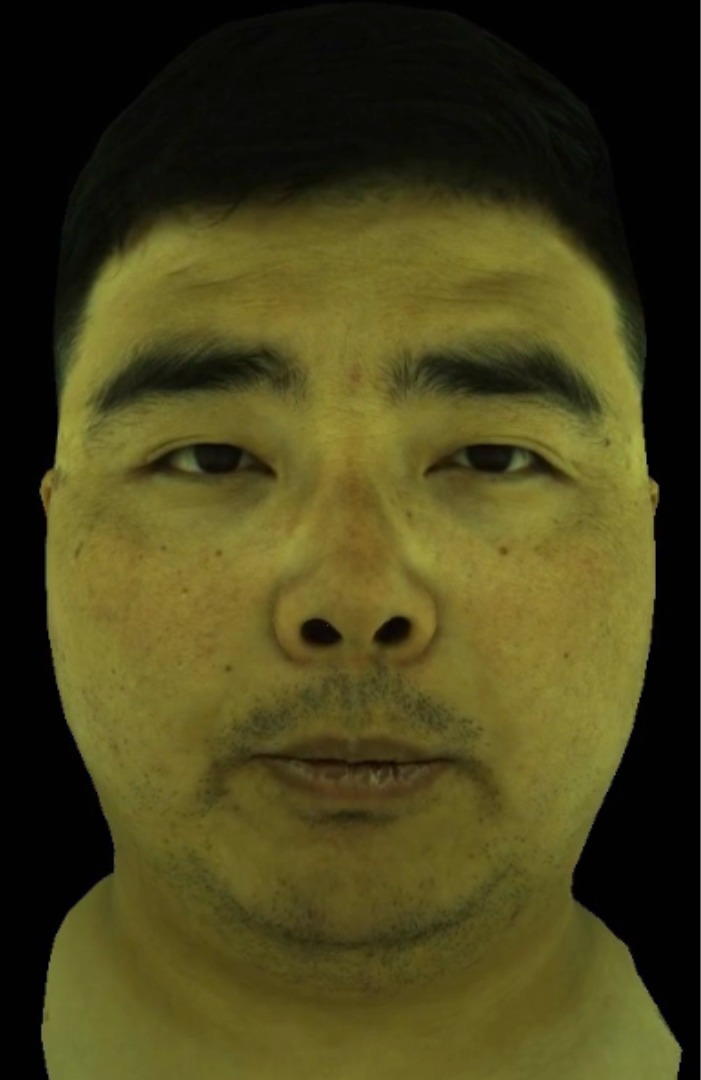}\\\hline
    \scriptsize{image} & \scriptsize{MCA} & \scriptsize{image} & \scriptsize{MCA} & \scriptsize{original} & \scriptsize{amplified} & \scriptsize{original} & \scriptsize{amplified}\\
    \specialrule{.1em}{.05em}{.05em}
    \end{tabu}
    \caption{\footnotesize Two new applications enabled by the MCA model. Left side shows two examples of flexible animation. Right side shows two examples of eye amplification.}
    \label{fig:app}
\end{figure}

\noindent
\textbf{Flexible animation:} Making funny expressions is part of social interaction. The MCA model can naturally better facilitate this task due to stronger expressiveness. To showcase this, we shuffle the head-mounted image sequences separately for each module, and randomly match them to simulate flexible expressions. It can be seen from Fig.~\ref{fig:app} that MCA produces natural flexible expressions, even though such expressions have never been seen holistically in the training set. 

\noindent
\textbf{Eye amplification:} In practical VR telepresence, we observe users often do not open their eyes to the full natural extend. This maybe due to muscle pressure from the headset wearing, and display light sources near the eyes. We introduce an eye amplification control knob to address this issue. In MCA, this can be simply accomplished by identifying the base $\mathbf{c}^{\mathrm{part}}_k$ that correspond to closed eye, and amplifying the latent space distance by multiplying a user-provided amplification magnitude. Fig.~\ref{fig:app} shows examples of amplifying by a factor of 2.

%% file: 6_conclusion.tex
\section{Conclusion}
We addressed the problem of VR telepresence, which aimed to provide remote and immersive face-to-face telecommunication through VR headsets. 
Codec Avatar (CA) utilized view-dependent neural networks to achieve realistic facial animation. 
We presented a new formulation of codec avatar named Modular Codec Avatar (MCA). 
This paper combines classic module-based face modeling with codec avatars in VR telepresence.
We presented several important techniques to realize MCA effectively.
We demonstrated that MCA achieves improved expressiveness and robustness through experiments on a comprehensive real-world dataset that emulated practical scenarios.
New applications in VR telepresence enabled by the proposed model were finally showcased.